\documentclass[10pt,journal,compsoc,twoside]{IEEEtran}
\usepackage[accsupp]{axessibility}  
\usepackage{times}
\usepackage{epsfig}
\usepackage{graphicx}
\usepackage{amsmath}
\usepackage{amssymb}

\usepackage{subcaption}
\usepackage{mathtools}
\usepackage{bbm}
\usepackage{colortbl}
\usepackage[flushleft]{threeparttable}
\usepackage{booktabs}
\usepackage{multirow}
\usepackage{color}
\usepackage{xcolor}
\usepackage{textcomp}  
\usepackage{gensymb}
\usepackage[export]{adjustbox}
\usepackage{tabulary}
\newcolumntype{x}[1]{>{\centering\arraybackslash}p{#1pt}}
\usepackage{pifont}
\usepackage{pgfplots}
\pgfplotsset{compat=1.18}

\usepackage{MnSymbol}
\usepackage{footmisc}
\usepackage{extramarks}

\newcommand{\cmark}{\ding{51}}%
\newcommand{\xmark}{\ding{55}}%

\definecolor{citecolor}{RGB}{34,139,34}  
\usepackage[pagebackref=true,breaklinks=true,colorlinks,citecolor=citecolor,bookmarks=false]{hyperref}

\usepackage [misc]{ifsym}
\newcommand{\corrAuthor}{$^{\textrm{\Letter}}$}

\definecolor{ygcolor}{RGB}{0,50,160}

\definecolor{xycolor}{RGB}{0,150,150}

\definecolor{rucolor}{RGB}{201,7,22}

\definecolor{RoneColorBG}{rgb}{1, 1, 1}
\definecolor{RtwoColorBG}{rgb}{1, 1, 1}
\definecolor{RthreeColorBG}{rgb}{1, 1, 1}
\definecolor{RoneColor}{rgb}{0, 0, 0}
\definecolor{RtwoColor}{rgb}{0, 0, 0}
\definecolor{RthreeColor}{rgb}{0, 0, 0}

\newcommand{\Rone}[1]{\textcolor{RoneColor}{#1}}
\newcommand{\Rtwo}[1]{\textcolor{RtwoColor}{#1}}
\newcommand{\Rthree}[1]{\textcolor{RthreeColor}{#1}}

\DeclareMathOperator{\avg}{avg}

\DeclareCaptionLabelFormat{andtable}{#1~#2  \&  \tablename~\thetable}

\newcommand{\etal}[0]{\textit{et al.}}
\newcommand{\ie}[0]{\textit{i.e.}}
\newcommand{\eg}[0]{\textit{e.g.}}
\newcommand{\wrt}[0]{\textit{w.r.t.~}}
\newcommand{\vs}[0]{\textit{v.s.~}}

\newcommand{\loss}[0]{\mathcal{L}}
\newcommand{\pose}[0]{\mathbf{P}}

\newcommand{\rot}[0]{\mathbf{R}}
\newcommand{\trans}[0]{\mathbf{t}}

\newcommand{\obj}[0]{\mathcal{O}}
\newcommand{\model}[0]{\mathcal{M}}
\newcommand{\img}[0]{\mathbf{I}}
\newcommand{\depth}[0]{\mathbf{D}}
\newcommand{\coor}[0]{\mathbf{C}}
\newcommand{\point}[0]{\mathbf{x}}
\newcommand{\mask}[0]{\mathbf{M}}
\newcommand{\Msra}[0]{\mathbf{M}_\text{SRA}}
\newcommand{\Msragt}[0]{\mathbf{\bar{M}}_\text{SRA}}
\newcommand{\Msraest}[0]{\mathbf{\hat{M}}_\text{SRA}}
\newcommand{\Mcorr}[0]{\mathbf{M}_\text{2D-3D}}
\newcommand{\Mthreed}[0]{\mathbf{M}_\text{3D}}
\newcommand{\Mvis}[0]{\mathbf{M}_\text{vis}}
\newcommand{\Mvisgt}[0]{\mathbf{\bar{M}}_\text{vis}}
\newcommand{\Mvisest}[0]{\mathbf{\hat{M}}_\text{vis}}
\newcommand{\Mfull}[0]{\mathbf{M}_\text{amodal}}
\newcommand{\Mfullgt}[0]{\mathbf{\bar{M}}_\text{amodal}}
\newcommand{\Mfullest}[0]{\mathbf{\hat{M}}_\text{amodal}}
\newcommand{\Mxyz}[0]{\mathbf{M}_\text{XYZ}}
\newcommand{\Mxyzgt}[0]{\mathbf{\bar{M}}_\text{XYZ}}
\newcommand{\Mxyzest}[0]{\mathbf{\hat{M}}_\text{XYZ}}

\newcommand{\app}{\raise.17ex\hbox{$\scriptstyle\sim$}}
\def\x{\times}

\newlength\savewidth\newcommand\shline{\noalign{\global\savewidth\arrayrulewidth
  \global\arrayrulewidth 1pt}\hline\noalign{\global\arrayrulewidth\savewidth}}
\newcommand{\tablestyle}[2]{\setlength{\tabcolsep}{#1}\renewcommand{\arraystretch}{#2}\centering\footnotesize}

\newcommand\Set[2]{\{\,#1\mid#2\,\}}

\newcommand\copyrighttext{%
  \footnotesize \textcopyright 2025 IEEE. Personal use of this material is permitted.  
  Permission from IEEE must be obtained for all other uses, in any current or future media, including reprinting/republishing this material for advertising or promotional purposes, creating new collective works, for resale or redistribution to servers or lists, or reuse of any copyrighted component of this work in other works.
  DOI: \href{https://doi.org/10.1109/TPAMI.2025.3553485}{10.1109/TPAMI.2025.3553485}
  }
\newcommand\copyrightnotice{%
\begin{tikzpicture}[remember picture,overlay]
\node[anchor=south,yshift=5pt] at (current page.south)
{\fbox{\parbox{\dimexpr\textwidth-\fboxsep-\fboxrule\relax}{\copyrighttext}}};
\end{tikzpicture}%
}

\begin{document}

\title{GDRNPP: A Geometry-guided and Fully Learning-based Object Pose Estimator}
\markboth{IEEE TRANSACTIONS ON PATTERN ANALYSIS AND MACHINE INTELLIGENCE}%
{Liu \MakeLowercase{\textit{et al.}}: GDRNPP: A Geometry-guided and Fully Learning-based Object Pose Estimator}
    
\author{Xingyu Liu$^{\dag}$,
        Ruida Zhang$^{\dag}$,
        Chenyangguang Zhang,
        Gu Wang\corrAuthor,\\
        Jiwen Tang,
        Zhigang Li,
        and Xiangyang Ji,~\IEEEmembership{Member,~IEEE} \corrAuthor
\IEEEcompsocitemizethanks{
\IEEEcompsocthanksitem 
Xingyu Liu, Ruida Zhang, Chenyangguang Zhang, Zhigang Li, and Xiangyang Ji are with the Department of Automation, Tsinghua University, Beijing 100084, China, and also with BNRist, Beijing 100084, China.
E-mail: \{liuxy21,zhangrd23,zcyg22\}@mails.tsinghua.edu.cn, lzg.matrix@gmail.com, xyji@tsinghua.edu.cn.
\IEEEcompsocthanksitem Gu Wang is with the Lab for High Technolodgy, Tsinghua University, Beijing 100084, China.
E-mail: guwang12@gmail.com.
\IEEEcompsocthanksitem Jiwen Tang is with the School of Information Engineering, China University of Geosciences Beijing, Beijing 100084, China.
E-mail: Rainbowend@163.com.
}
\thanks{\dag: Xingyu Liu and Ruida Zhang have equally contributed.}
\thanks{\corrAuthor: Corresponding authors.}
}

\IEEEtitleabstractindextext{%
\begin{abstract}
6D pose estimation of rigid objects is a long-standing and challenging task in computer vision.
Recently, the emergence of deep learning reveals the potential of Convolutional Neural Networks (CNNs) to predict reliable 6D poses.
Given that direct pose regression networks currently exhibit suboptimal performance, most methods still resort to traditional techniques to varying degrees.
For example, top-performing methods often adopt an indirect strategy by first establishing 2D-3D or 3D-3D correspondences followed by applying the RANSAC-based P$n$P or Kabsch algorithms, and further employing ICP for refinement.
Despite the performance enhancement, the integration of traditional techniques makes the networks time-consuming and not end-to-end trainable. 
Orthogonal to them, this paper introduces a fully learning-based object pose estimator. 
In this work, we first perform an in-depth investigation of both direct and indirect methods and propose a simple yet effective Geometry-guided Direct Regression Network (GDRN) to learn the 6D pose from monocular images in an end-to-end manner.
Afterwards, we introduce a geometry-guided pose refinement module, enhancing pose accuracy when extra depth data is available.
Guided by the predicted coordinate map, we build an end-to-end differentiable architecture that establishes robust and accurate 3D-3D correspondences between the observed and rendered RGB-D images to refine the pose.
\Rtwo{Our enhanced pose estimation pipeline GDRNPP (GDRN Plus Plus) conquered the leaderboard of the BOP Challenge for two consecutive years, becoming the first to surpass all prior methods that relied on traditional techniques in both accuracy and speed.}
The code and models are available at \href{https://github.com/shanice-l/gdrnpp_bop2022}{https://github.com/shanice-l/gdrnpp\_bop2022}.
\end{abstract}
}

\maketitle
\copyrightnotice

\IEEEdisplaynontitleabstractindextext

\IEEEpeerreviewmaketitle

\IEEEraisesectionheading{\section{Introduction}\label{sec:introduction}}

\IEEEPARstart{E}{stimating} the 6D pose, \ie, the 3D rotation and 3D translation, of objects in the camera frame is a fundamental problem in computer vision.
It has wide applicability to many real-world tasks such as robotic manipulation~\cite{collet2011moped,zhu2014single,tremblay2018deep}, augmented reality~\cite{marchand2015pose,tang20193d} and autonomous driving~\cite{manhardt2019roi,Wu_2019_CVPRW_6dvnet}.
In the pre-deep learning era, methods can be roughly categorized into feature-based~\cite{guo20143d,aldoma2011cad,rusu2010fast} and template-based~\cite{hinterstoisser2010dominant,hinterstoisser2011multimodal,Hinterstoisser2012} approaches.
Among these, the most representative branch of work is based on point pair features (PPFs), which is proposed by Drost \etal~\cite{drost2010model} and still achieves competitive results in recent years~\cite{vidal2018method}.
Nonetheless, with the advent of deep learning, methods based on neural networks become dominant in \Rthree{instance-level} object pose estimation~\cite{brachmann2016uncertainty,peng2019pvnet,labbe2020CosyPose,wang2019densefusion,he2020pvn3d,he2021ffb6d}.

\Rthree{Given the CAD model of objects}, different strategies for predicting 6D pose from monocular or depth data have been proposed.
An intuitive approach is to directly regress 6D poses from neural networks~\cite{manhardt2019ambiguity,wang2019densefusion,jiang2022uni6d}.
Unfortunately, due to the lack of geometric prior, \Rthree{such as 2D-3D or 3D-3D correspondences}, these methods currently exhibit suboptimal performance when compared with approaches that instead rely on establishing 2D-3D~\cite{li2019cdpn,hodan2020epos} or 3D-3D correspondences~\cite{he2020pvn3d,he2021ffb6d,shugurov2021dpodv2} to estimating the 6D pose.

Differently, this latter class of methods usually involves solving the 6D pose through traditional techniques like P$n$P or Kabsch,
and they oftentimes employ Iterative Closest Point (ICP) algorithm for further depth refinement.
While such a paradigm provides good estimates, it also suffers from several drawbacks.
First, these methods are usually trained with a surrogate objective for correspondence regression, which does not necessarily reflect the actual 6D pose error after optimization.
In practice, two sets of correspondences can have the same average error while describing completely different poses.
Second, correspondence-based methods are sensitive to outliers, rendering the algorithms not robust and prone to being trapped in local minima.
Therefore, they often resort to non-differentiable filtering algorithms like RANSAC, which limits their applicability in tasks requiring differentiable poses. 
For instance, these methods cannot be coupled with self-supervised learning from unlabeled real data~\cite{wang2020self6d,manhardt2020cpspp,beker2020monocular_self3d_det,wang2021occlusion} \Rthree{or joint optimization of 3D reconstruction and poses for scene understanding \cite{Zhang_2021_CVPR}}, 
as they require the computation of the pose to be fully differentiable in order to obtain a signal between data and pose.
Besides, the whole process can be very time-consuming when dealing with dense correspondences.

\begin{figure}[t]
    \begin{center}
    \includegraphics[width=0.8\linewidth]{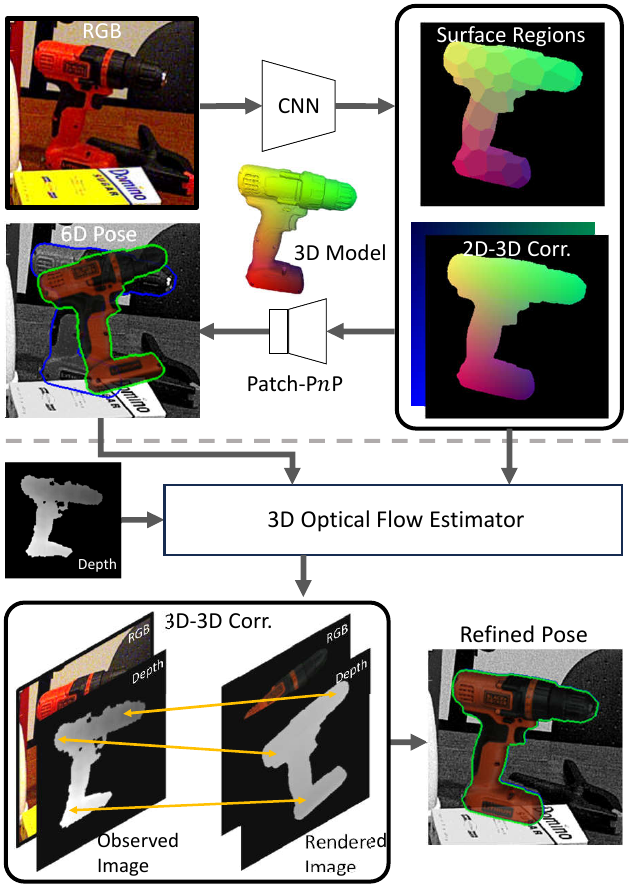}
    \end{center}
    \caption{
        {\bf Illustration of GDRNPP.}
        Firstly, we directly regress the 6D object pose from a single RGB using a CNN and the learnable Patch-P$n$P by leveraging the guidance of intermediate geometric features including 2D-3D dense correspondences and surface region attention.
        Moreover, when depth information is available, the network predicts the 3D optical flow to establish 3D-3D correspondences between the observed and rendered RGB-D image to refine the pose. \Rthree{The details are elaborated in Fig.~\ref{fig:framework} and Fig.~\ref{fig:refinement_pipeline}.}
    }
    \label{fig:teaser}
\end{figure}

To summarize, while correspondence-based methods currently dominate the field, 
the incorporation of traditional techniques renders the pipelines time-consuming and non-end-to-end trainable.
To tackle this problem, we seek to build a geometry-guided and fully learning-based object pose estimator in this work, as illustrated in Fig.~\ref{fig:teaser}.

Firstly, to circumvent the non-differentiable and lengthy P$n$P/RANSAC process, our network establishes 2D-3D correspondences whilst computing the final 6D pose estimate in a fully differentiable way. 
In its core, we propose to learn the P$n$P optimization from intermediate geometric representations, exploiting the fact that the correspondences are organized in image space, 
which gives a significant boost in performance, outperforming all prior monocular-based works.

Additionally, when depth information is accessible, we extend our pipeline to incorporate the extra modality by introducing a trainable geometry-guided pose refinement module.
Drawing inspiration from \cite{lipson2022coupled}, we adopt the ``render and compare'' strategy and predict the 3D optical flow between the rendered image and observed image to establish 3D-3D dense correspondences to solve the pose.
\Rone{
Previous methods \cite{hu2022perspective, lipson2022coupled} mostly rely on RGB images to estimate optical flow.
While effective in many cases, these methods face limitations when there are significant discrepancies between the rendered and observed images, such as variations in lighting conditions or object materials.
To address this, our approach incorporates domain-invariant coordinates 
as an additional input, enhancing robustness and mitigating such challenges when they arise.}
Thanks to the learning-based refinement module and the domain-invariant information in the coordinate map, the correspondences are robust and accurate without relying on the traditional non-differentiable filtering method like RANSAC, thus leading to a substantial performance boost.

The overall pipeline, which we dub GDRNPP \Rtwo{(GDRN Plus Plus)}, offers a flexible framework that adapts to the availability of either RGB or depth modality, ensuring accurate and robust 6D pose estimation.
To sum up, our technical contributions are threefold:
\begin{itemize}
    \item We construct a fully learning-based object pose estimation pipeline, \Rtwo{achieving state-of-the-art performance among existing 6D pose estimation methods in both RGB and RGB-D settings.} 
    \item We propose a simple yet effective Geometry-guided Direct Regression Network (GDRN) to boost the performance of monocular-based 6D pose estimation by leveraging the geometric guidance from dense correspondence-based features.
    \item We further devise a geometry-guided refinement module, 
    enhancing pose accuracy when extra depth data is accessible.
    The predicted object coordinates are leveraged to set up more elaborated 3D-3D dense correspondences between the observed and rendered RGB-D images, leading to more precise pose estimation.
\end{itemize}
\Rtwo{Notably, GDRNPP conqured the leaderboard on the Benchmark for 6D Object Pose Estimation (BOP) Challenge in 2022 and 2023~\cite{sundermeyer2023bop,hodan2024bop}, winning most of pose and detection awards.
The whole pipeline was recognized as ``The Overall Best Method'' for two consecutive years.}
For the first time in the BOP Challenge, the deep-learning-based method distinctly \Rtwo{surpassed} traditional methods leveraging PPFs or ICP in both accuracy and speed.

Compared to the former version of this work (GDR-Net) published in CVPR 2021~\cite{wang2021gdr}, the revised GDRNPP makes the following improvements.
First, we conduct a series of exploratory analyses to strengthen GDRN, including more accurate detection, improved augmentation and enhanced model architecture, yielding substantial improvements to our baseline.
Second, we devise a geometry-guided pose refinement module that predicts 3D-3D dense correspondences between the observed and rendered images to refine the pose when depth is available.
The refinement procedure not only boosts performance but also raises the versatility of our pipeline, enabling it to flexibly accommodate either RGB or RGB-D modalities.
Moreover, in contrast to \cite{wang2021gdr}, GDRNPP demonstrates enhanced capability in generating reliable poses in challenging circumstances, especially with the T-LESS and ITODD datasets characterized by numerous symmetric objects with a conspicuous absence of texture.

\section{Related Work}
In this section, we review some prominent pioneer works in the field of 6D pose estimation.
These works can be roughly divided into three categories which are indirect methods, direct methods and differentiable indirect methods.
Subsequently, we introduce several commonly employed strategies for pose refinement.

\subsection{Indirect Methods}
The most popular approach is to establish 2D-3D or 3D-3D correspondences, which are then leveraged to solve for the 6D pose using a variant of the RANSAC-based P$n$P/Kabsch algorithm.
For instance, BB8~\cite{rad2017bb8} and YOLO6D~\cite{tekin18_yolo6d} compute the 2D projections of a set of fixed control points (\eg the 3D corners of the encapsulating bounding box).
To enhance the robustness, 
PVNet~\cite{peng2019pvnet} additionally conducts segmentation coupled with voting for each correspondence.
HybridPose~\cite{song2020hybridpose} extends PVNet by predicting edges and axes of symmetries at the same time.
\Rtwo{K\"onig \textit{et al.}~\cite{konig2020hybrid} develop a fast point pair voting approach for improvement of efficiency.}
Moreover, PVN3D~\cite{he2020pvn3d} extends the idea of keypoint voting to 3D space, leveraging a deep Hough voting network to detect 3D keypoints, 
\Rtwo{while RCVPose~\cite{wu2022vote} devises a radial keypoint voting strategy to improve voting accuracy.}
Meanwhile, FFB6D~\cite{he2021ffb6d} works on the fusion of color and depth features, introducing a full flow bidirectional fusion network for 3D keypoints prediction.
However, the recent trend goes towards predicting dense rather than sparse correspondences, including DPOD~\cite{zakharov2019dpod}, DPODv2~\cite{shugurov2021dpodv2}, CDPN~\cite{li2019cdpn}, SurfEmb~\cite{haugaard2022surfemb}, and SDFlabel~\cite{zakharov2020autolabeling}.
They follow the assumption that a larger number of correspondences will mitigate the problem of their inaccuracies and will result in more precise poses.
\Rtwo{There are also effective endeavors developed in order to construct more robust dense correspondences.} 
Pixel2Pose~\cite{park2019pix2pose} leverages a GAN on top of dense correspondences to increase stability. 
EPOS~\cite{hodan2020epos} makes use of fragments in order to account for ambiguities in pose.
\Rtwo{Recently, ZebraPose~\cite{su2022zebrapose} leverages a binary surface code for enhanced efficiency to set up 2D-3D correspondences in a coarse-to-fine manner.}
\Rtwo{Compared to the aforementioned methods, GDRN predicts intermediate
geometric features including 2D-3D dense correspondences, meanwhile differentiably predicting the 6DoF pose.}

Another orthogonal line of work aims at learning a latent embedding of pose which can be utilized for retrieval during inference.
These embeddings are commonly either grounded on metric learning employing a triplet loss~\cite{wohlhart2015learning},
or via training of an Auto-Encoder~\cite{sundermeyer2018AAE,Sundermeyer_2020_CVPR, li2020pose}.

\subsection{Direct Methods}
Indirect methods leveraging correspondences have natural flaws in employing many tasks, which require the pose estimation to be differentiable~\cite{wang2020self6d,wang2021occlusion}.
Hence, some methods directly regress the 6D pose, either leveraging a point matching loss~\cite{xiang2017posecnn,cao2022dgecn}
or employing separate loss terms for each component~\cite{manhardt2019ambiguity,lienet_do2018,wang2019densefusion}.
Other methods discretize the pose space and conduct classification rather than regression~\cite{kehl2017ssd6d}.
A few methods also try to solve a proxy task during optimization.
Thereby, Manhardt \etal~\cite{manhardt2018deep} propose to employ an edge-alignment loss using the distance transform,
while Self6D~\cite{wang2020self6d} and Self6D++~\cite{wang2021occlusion} harness differentiable rendering to allow training on unlabeled samples.
Although direct regression methods seem simple and straightforward, they oftentimes perform worse than indirect methods due to the lack of 3D geometric knowledge.
Therefore, some methods attempt to eliminate this problem by introducing depth data.
For example, DGECN~\cite{cao2022dgecn} estimates depth and leverages it to guide the predictions of pose using an edge convolutional network from correspondences.
DenseFusion~\cite{wang2019densefusion} leverages CNN and PointNet~\cite{qi2017pointnet} separately to extract color and depth features and fuse them by matching each point, and further predicts pixel-wise poses with a neural network.
In contrast, Uni6D~\cite{jiang2022uni6d} direct concatenates RGB and depth with positional encoding and feeds them to an end-to-end network based on Mask-RCNN~\cite{he2017mask}.

GDR-Net~\cite{wang2021gdr}, the conference version of this paper, introduces a Patch-P$n$P module to replace P$n$P/RANSAC and make the monocular pose estimation pipeline differentiable.
Building on this concept, SO-Pose~\cite{Di2021SOPoseES} utilizes multiple geometry representations for 6D object pose estimation in scenes with occlusion or truncation.
Moreover, PPP-Net \cite{gao2022polarimetric} leverages polarized RGB images to effectively handle transparent or reflective objects.

\subsection{Differentiable Indirect Methods}

Recently, there has been an emerging trend of attempting to make P$n$P/RANSAC differentiable. 
In \cite{brachmann2018DSACpp}, \cite{brachmann2017dsac}, and \cite{brachmann2019NGRANSAC}, the authors introduce a novel differentiable way to apply RANSAC via sharing of hypotheses based on the predicted distribution. 
Nonetheless, these approaches require a complex training strategy, as they expect a good initialization for the scene coordinates. 
\Rthree{More recently, $\nabla$-RANSAC~\cite{wei2023generalized} proposes to learn inlier probabilities as an objective and incorporates Gumbel Softmax~\cite{jang2016categorical} relaxation to estimate gradients within the sampling distribution.}
As for P$n$P,
BP$n$P~\cite{chen2020BPnP} employs the Implicit Function Theorem~\cite{krantz2012IFT} to enable the computation of analytical gradients \wrt the pose loss. Yet, it is computationally expensive especially given too many correspondences since P$n$P/RANSAC is still needed for both training and inference. 
Instead, Single-Stage Pose~\cite{hu2020single_stage} attempts to learn the P$n$P stage with a PointNet-based architecture~\cite{qi2017pointnet} which learns to infer the 6D pose from a fixed set of sparse 2D-3D correspondences.
More recently, EPro-P$n$P~\cite{chen2022epro} makes the P$n$P layer differentiable by translating the output from the deterministic pose to a distribution of pose.

\subsection{Pose Refinement Methods}
\label{rw_pose_refine}
Several studies have delved into the realm of refinement methods to improve pose accuracy, as it is challenging to obtain accurate pose estimates in a single shot.
As for monocular methods, DeepIM~\cite{li2019deepim} is a representative approach that introduces the iterative ``render-and-compare'' strategy to CNN-based pose refinement. 
In each iteration, DeepIM renders the 3D model using the current pose estimate and then regresses a pose residual by comparing the rendered image with the observed image. 
Building upon this concept, CosyPose~\cite{labbe2020CosyPose} further leverages the multi-view information to match each individual objects and jointly refine a single global scene.
RePose~\cite{iwase2021repose} and RNNPose~\cite{xu2022rnnpose} formulate the pose refinement as an optimization problem based on feature alignment or the estimated correspondence field.

As for depth-based methods, the Iterative Closest Point (ICP) algorithm~\cite{ICP} and its variants~\cite{ICP1, ICP2, ICP3, ICP4, ICP5} stand out as the predominant traditional pose refinement algorithms. 
They have broad applications in monocular~\cite{labbe2020CosyPose,li2019cdpn} or depth~\cite{wang2019densefusion,he2020pvn3d,he2021ffb6d} based pose estimation methods.
Starting from an initial estimate, they repeatedly identify point-level correspondences and refine the pose based on these correspondences. 
However, due to the lack of prior knowledge of the object, the correspondences often contain multiple outliers and lead to the algorithm being trapped by local minima. 
More recently, learning-based methods adopt the ``render-and-compare" strategy to utilize the 3D model information of the objects to enhance the robustness of the correspondences.
\Rone{
In these methods, given an initial pose, a synthetic image and depth map are rendered based on the object's pose, then compared to the observed image to iteratively update the pose until convergence.
For example, $se(3)$-TrackNet~\cite{wen2020se} utilizes two different networks to extract the features of the observed and rendered RGB-D images, and directly regresses the relative pose in $se(3)$.
Some approaches like PFA~\cite{hu2022perspective} predict 2D optical flow between the rendered and observed images to establish dense correspondences, thereby enhancing robustness.
However, a critical challenge arises when a corresponding point in one image does not precisely align with a pixel in the other image but instead falls between several pixels. 
In such cases, the depth value of the corresponding point must be interpolated, inevitably introducing errors due to discrepancies between the interpolated depth and the true depth. 
These errors can significantly impact pose estimation accuracy, especially near object edges, where interpolation can result in pronounced depth estimation errors.
To address these challenges, Coupled Iterative Refinement (CIR) \cite{lipson2022coupled} introduces a 3D optical flow estimator \cite{teed2020raft} that explicitly estimates the depth of corresponding points by leveraging RGB and depth information from both images. 
This approach enables more accurate depth computations, thereby enhancing pose estimation precision.
}
Inspired by \cite{lipson2022coupled}, we further utilize the predicted object coordinates from GDRN as prior knowledge to establish more accurate correspondences and enhance pose refinement.

\begin{figure*}[t]
    \begin{center}
    \includegraphics[width=0.96\linewidth]{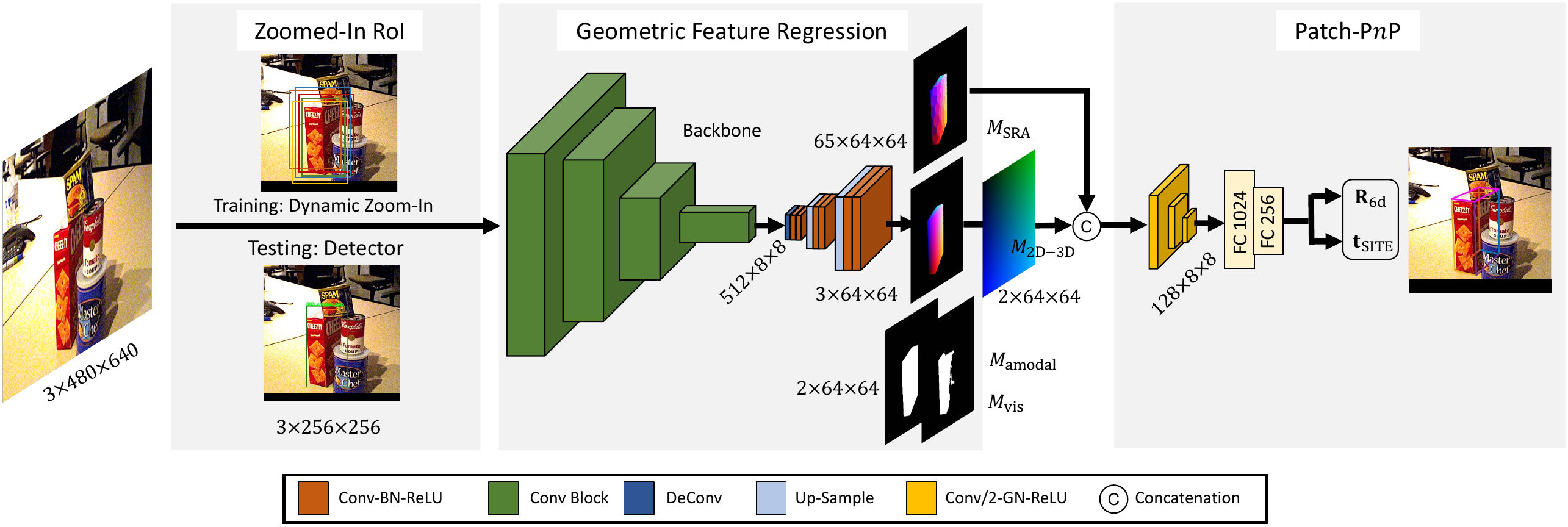}
    \end{center}
    \caption{
        \label{fig:framework}
        {\bf Framework of GDRN.}
        Given an RGB image $I$, our GDRN takes the zoomed-in RoI (Dynamic Zoom-In for training, off-the-shelf detections for testing) as input and predicts several intermediate geometric features.
        Then the Patch-P$n$P directly regresses the 6D object pose from \emph{Dense Correspondences} ($\Mcorr$) and \emph{Surface Region Attention} ($\Msra$).
    }

\end{figure*}

\section{Methods}

Given an RGB(-D) image $\img$ and a set of $L$ objects $\obj = \Set{\obj_i}{i=1,\cdots,L}$
together with their corresponding 3D CAD models $\model = \Set{\model_i}{i=1,\cdots,L}$,
our goal is to estimate the 6D pose $\pose=[\rot|\trans]$ \wrt the camera for each object present in $\img$.
Notice that $\rot$ describes the 3D rotation and $\trans$ denotes the 3D translation of the detected object.

Fig.~\ref{fig:framework} and \Rtwo{Fig.~\ref{fig:refinement_pipeline}} present a schematic overview of the proposed methodology.
In the core, we first detect all objects of interest using an off-the-shelf object detector,
such as~\cite{faster,tian2019fcos,ge2021yolox}.
For each detection, we then zoom in on the corresponding Region of Interest (RoI) and
feed it to our network to predict several intermediate geometric feature maps, \Rthree{\ie, dense correspondences maps and surface region attention maps}.
Thereby, we directly regress the associated 6D object pose from the \Rthree{intermediate geometric features}.
Additionally, 
when depth information is accessible, we predict the 3D optical flow between observed and rendered RGB-D images and build accurate and robust 3D-3D dense correspondences to refine the pose.

In the following, we first (Sec.~\ref{sec:revisit}) revisit the key ingredients of direct 6D object pose estimation methods.
Afterwards (Sec.~\ref{sec:GDRN}), we illustrate a simple yet effective Geometry-Guided Direct Regression Network (GDRN)
which unifies regression-based \emph{direct} methods and geometry-based \emph{indirect} methods,
thus harnessing the best of both worlds.
Finally (Sec.~\ref{sec:refine}), we introduce the geometry-guided pose refinement module which leverages depth information to further boost the accuracy.

\subsection{Revisiting Direct 6D Object Pose Estimation}
\label{sec:revisit}
Direct 6D pose estimation methods usually differ in one or more of the following components.
Firstly, the parameterization of the rotation $\rot$ and translation $\trans$, and
secondly, the employed loss for pose.
In this section, we investigate different commonly used parameterizations and demonstrate that
appropriate choices have a significant impact on the 6D pose estimates.

\noindent \textbf{Parameterization of 3D Rotation.}
Several different parameterizations can be employed to describe 3D rotations.
Since many representations exhibit ambiguities, \ie~$\rot_i$ and $\rot_j$ describe the same rotation with $\rot_i \neq \rot_j$,
most works rely on parametrizations that are unique to help training.
Therefore, common choices are unit quaternions~\cite{xiang2017posecnn,manhardt2018deep,li2019deepim},
log quaternions~\cite{park2020latentfusion}, or Lie algebra-based vectors~\cite{lienet_do2018}.

Nevertheless, it is well-known that all representations with four or fewer dimensions for 3D rotation have discontinuities in Euclidean space.
When regressing a rotation, this introduces an error close to the discontinuities which becomes often significantly large.
To overcome this limitation, ~\cite{rot6d_zhou2019} proposed a novel continuous 6-dimensional
representation for $\rot$ in $SO(3)$, which has proven promising~\cite{rot6d_zhou2019,labbe2020CosyPose}.
Specifically, the 6-dimensional representation $\rot_{\text{6d}}$ is defined as the first two columns of $\rot$
\begin{equation}
\label{eq:r6}
    \rot_{\text{6d}} = \left[\rot_{\boldsymbol{\cdot}1} ~|~ \rot_{\boldsymbol{\cdot}2}\right].
\end{equation}
Given a 6-dimensional vector $\rot_{\text{6d}} = [\mathbf{r}_1 | \mathbf{r}_2]$,
the rotation matrix $\rot = [\rot_{\boldsymbol{\cdot}1} | \rot_{\boldsymbol{\cdot}2} | \rot_{\boldsymbol{\cdot}3}]$
can be computed according to
\begin{equation}
\begin{cases}
\rot_{\boldsymbol{\cdot}1} = \phi(\mathbf{r}_1) \\
\rot_{\boldsymbol{\cdot}3} = \phi(\rot_{\boldsymbol{\cdot}1} \times \mathbf{r}_2) \\
\rot_{\boldsymbol{\cdot}2} = \rot_{\boldsymbol{\cdot}3} \times \rot_{\boldsymbol{\cdot}1} \\
\end{cases},
\label{eq:r6_to_rot}
\end{equation}
where $\phi(\bullet)$ denotes the vector normalization operation.

Given the advantages of this representation, in this work we employ $\rot_{\text{6d}}$ to parameterize the 3D rotation.
Nevertheless, in contrast to \cite{rot6d_zhou2019,labbe2020CosyPose}, we propose to let the network predict the
allocentric representation~\cite{kundu20183d} of rotation. 
This representation is favored as it is viewpoint-invariant under 3D translations of the object.
Hence, it is more suitable to deal with zoomed-in RoIs.
Note that the egocentric rotation can be easily converted from allocentric rotation given 3D translation and
camera intrinsics $\mathbf{K}$ following~\cite{kundu20183d}.

\noindent \textbf{Parameterization of 3D Translation.}
Since directly regressing the translation $\trans=[t_x, t_y, t_z]^\top \in \mathbb{R}^3$
in 3D space does not work well in practice, previous works usually decouple the translation into the
2D location  $(o_x, o_y)$ of the projected 3D centroid and the object's distance $t_z$ towards the camera.
Given the camera intrinsics $\mathbf{K}$, the translation can be calculated via back-projection
\begin{equation}
    \trans = \mathbf{K}^{-1} t_z\left[o_x, o_y, 1\right]^\top.
\label{eq:t_bp}
\end{equation}
Exemplary, \cite{kehl2017ssd6d,sundermeyer2018AAE} approximate $(o_x, o_y)$
as the bounding box center $(c_x, c_y)$ and estimate $t_z$ using a reference camera distance.
PoseCNN~\cite{xiang2017posecnn} directly regresses $(o_x, o_y)$ and $t_z$.
Nonetheless, this is not suitable for dealing with zoomed-in RoIs, since it is essential for
the network to estimate position and scale invariant parameters.

Therefore, in our work we utilize a Scale-Invariant representation for Translation Estimation (SITE)~\cite{li2019cdpn}.
Concretely, given the size $s_o=\max\{w, h\}$ and center $(c_x, c_y)$ of the detected bounding box and the ratio $r=s_\text{zoom}/s_o$ \wrt the zoom-in size $s_\text{zoom}$,
the network regresses the scale-invariant translation parameters $\trans_{\text{SITE}} = [\delta_x, \delta_y, \delta_z]^\top$, where
\begin{equation}
\begin{cases}
\delta_x =  (o_x - c_x) / w \\
\delta_y =  (o_y - c_y) / h \\
\delta_z =  t_z / r \\
\end{cases}.
\label{eq:t_site}
\end{equation}
Finally, the 3D translation can be solved according to Eq.~\ref{eq:t_bp}.

\noindent \textbf{Disentangled 6D Pose Loss.}
Apart from the parameterization of rotation and translation, the choice of loss function is also crucial for 6D pose optimization.
Instead of directly utilizing distances based on rotation and translation (\eg, angular distance, $L_1$ or $L_2$ distances),
most works employ a variant of Point-Matching loss~\cite{li2019deepim,xiang2017posecnn,labbe2020CosyPose} based on the
ADD(-S) metric~\cite{Hinterstoisser2012,Brachmann2014Learning6O} in an effort to couple the estimation of rotation and translation.

Inspired by ~\cite{Simonelli2019_disentangle_loss,labbe2020CosyPose}, we employ a novel variant of disentangled
6D pose loss via individually supervising
the rotation $\rot$, the scale-invariant 2D object center $(\delta_x, \delta_y)$, and the distance $\delta_z$.
\begin{equation}
    \loss_{\text{Pose}} = \loss_\rot + \loss_{\text{center}} + \loss_{z}.
\label{eq:loss_pose}
\end{equation}
Thereby,
\begin{equation}
    \begin{cases}
        \loss_\rot &= \underset{\mathbf{x} \in \mathcal{M}}{\avg} \| \hat{\rot} \mathbf{x} - \bar{\rot} \mathbf{x} \|_1 \\
        \loss_{\text{center}} &= \| (\hat{\delta}_x - \bar{\delta}_x,  \hat{\delta}_y - \bar{\delta}_y) \|_1 \\
        \loss_{z} &= \| \hat{\delta}_z - \bar{\delta}_z \|_1
    \end{cases},
\label{eq:loss_pose_detail}
\end{equation}
where $\hat{\bullet}$ and $\bar{\bullet}$ denote prediction and ground truth, respectively.
To account for symmetric objects,
given $\mathcal{\bar{\mathcal{R}}}$, the set of all possible ground-truth rotations under symmetry,
we further extend our loss to a symmetry-aware formulation $\loss_{\rot,\text{sym}} = \underset{\mathbf{\bar{\rot}} \in \mathcal{\bar{\mathcal{R}}}}{\min} \loss_\rot(\hat{\rot}, \bar{\rot})$.

\subsection{Geometry-guided Direct Regression Network}
\label{sec:GDRN}

In this section, we present our Geometry-guided Direct Regression Network, which we dub GDRN. Harnessing dense correspondence-based geometric features, we directly regress 6D object pose.
Thereby, GDRN unifies approaches based on dense correspondences and direct regression.

\noindent \textbf{Network Architecture.}
As shown in Fig.~\ref{fig:framework}, we feed the GDRN with a zoomed-in RoI of size $256\times256$ and predict three intermediate geometric feature maps with the spatial size of $64\times64$,
which are composed of the \emph{Dense Correspondences Map} ($\Mcorr$), the \emph{Surface Region Attention Map} ($\Msra$) and the \emph{Visible Object Mask} ($\Mvis$).
Especially, for heavily obstructed datasets, we additionally predict the full \emph{Amodal Object Mask} ($\Mfull$) to improve the capability to reason about occlusions.

Our network is inspired by CDPN~\cite{li2019cdpn}, a state-of-the-art dense correspondence-based method for indirect pose estimation.
In essence, we keep the layers
 for regressing $\Mxyz$ and $\Mvis$, while removing the disentangled translation head.
Additionally, we append the channels required by $\Msra$ to the output layer.
Since these intermediate geometric feature maps are all organized 2D-3D correspondences \wrt the image,
we employ a simple yet effective 2D convolutional \emph{Patch-P$n$P} module to directly
regress the 6D object pose from $\Mcorr$ and $\Msra$.

The Patch-P$n$P module consists of three convolutional layers with kernel size 3$\times$3 and $\text{stride}=2$,
each followed by Group Normalization~\cite{wu2018group} and ReLU activation.
Two Fully Connected (FC) layers are then applied to the flattened feature, reducing the dimension from 8192 to 256.
Finally, two parallel FC layers output the 3D rotation $\rot$ parameterized as $\rot_{\text{6d}}$ (Eq.~\ref{eq:r6}) and
3D translation $\trans$ parameterized as $\trans_\text{SITE}$ (Eq.~\ref{eq:t_site}), respectively.

\noindent \textbf{Dense Correspondences Maps ($\Mcorr$).}
In order to compute the Dense Correspondences Maps $\Mcorr$, we first estimate the underlying Dense Coordinates Maps ($\Mxyz$). $\Mcorr$ can then be derived by stacking $\Mxyz$ onto the corresponding 2D pixel coordinates.
In particular, given the CAD model of an object, $\Mxyz$ can be obtained by rendering the model's 3D object coordinates given the associated pose.
Similar to~\cite{li2019cdpn,wang2019nocs}, we let the network predict a normalized representation of $\Mxyz$.
Concretely, each channel of $\Mxyz$ is normalized within $[0,1]$ by $(l_x, l_y, l_z)$, which is the size of the corresponding tight 3D bounding box of the CAD model.

Notice that $\Mcorr$ does not only encode the 2D-3D correspondences, but also explicitly reflects the geometric shape information of objects.
Moreover, as previously mentioned, since $\Mcorr$ is regular \wrt the image, we are capable of learning the 6D object pose via a simple 2D convolutional neural network (Patch-P$n$P).

\begin{figure*}[t]
    \begin{center}
    \includegraphics[width=0.96\linewidth]{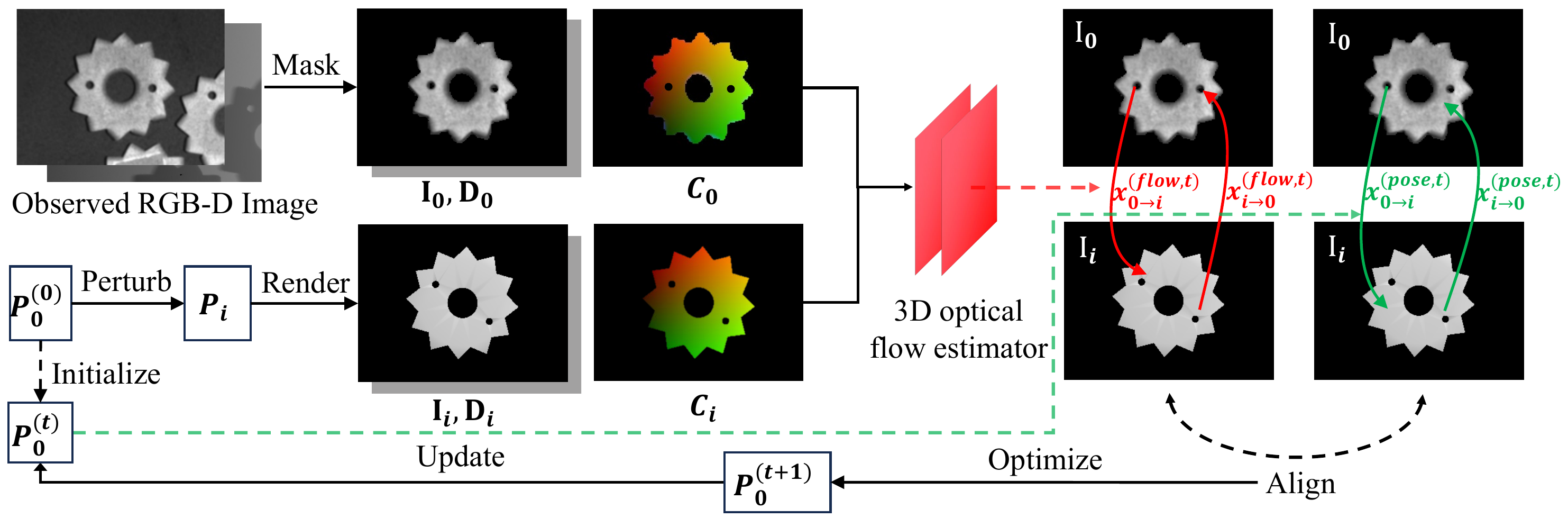}
    \end{center}

    \caption{
    \label{fig:refinement_pipeline}
    \Rtwo{
    {\bf Framework of the Refinement Module.}
    Starting with an initial pose $P_0^{(0)}$, perturbations are applied to generate a set of object poses $\{P_i \, | \, i = 1, 2, \dots, n\}$. 
    Correspondences between the observed image $I_0$ and the rendered images $\{I_i\}$ are established in two parallel ways: 
    (1) using a coordinate-guided 3D optical flow estimator to obtain $x_{0 \rightarrow i}^{(flow, t)}$ and $x_{i \rightarrow 0}^{(flow, t)}$, and 
    (2) using the predicted pose to derive $x_{0 \rightarrow i}^{(pose, t)}$ and $x_{i \rightarrow 0}^{(pose, t)}$. 
    By aligning these correspondences, the pose $P_0$ is iteratively refined, updating $P_0^{(t)}$ to $P_0^{(t+1)}$. 
    This optimization is repeated for $T=10$ iterations (inner loop), after which a new set of poses $\{P_i\}$ is generated, and the corresponding images are rendered. 
    The entire process is repeated $N_\text{out}=4$ times (outer loop) to achieve the final result.}
}

\end{figure*}

\noindent \textbf{Surface Region Attention Maps ($\Msra$).}
Inspired by~\cite{hodan2020epos},
we let the network predict the surface regions as additional ambiguity-aware supervision.
However, instead of coupling them with RANSAC, we use them within our Patch-P$n$P framework.

Essentially, the ground-truth regions $\Msra$ can be derived from $\Mxyz$ employing farthest points sampling.

For each pixel we classify the corresponding regions,
thus the probabilities in the predicted $\Msra$ implicitly represent the symmetry of an object.
For instance, if a pixel is assigned to two potential fragments due to a plane of symmetry,
minimizing this assignment will return a probability of 0.5 for each fragment. \Rthree{Therefore, the probability distribution of $\Msra$ reflect the symmetries of objects.}
Moreover, leveraging $\Msra$ not only mitigates the influence of ambiguities but also acts as an auxiliary task on top of $\Mthreed$. 
In other words, it eases the learning of $\Mthreed$ by first locating coarse regions and then regressing finer coordinates.
We utilize $\Msra$ as symmetry-aware attention \Rthree{input} to guide the learning of Patch-P$n$P.

\noindent \textbf{Geometry-guided 6D Object Pose Regression.}
The presented image-based geometric feature patches, \ie, $\Mcorr$ and $\Msra$,
are then utilized to guide our proposed Patch-P$n$P for direct 6D object pose regression as
\begin{equation}
    \label{eq:patch_pnp}
    \pose = \text{Patch-P}n\text{P}(\Mcorr, \Msra).
\end{equation}
We employ $\mathcal{L}_1$ loss for normalized $\Mxyz$, visible masks $\Mvis$ and amodal masks $\Mfull$, and cross-entropy loss ($CE$) for $\Msra$.
\begin{equation}
    \begin{split}
    \loss_{\text{Geom}} =~& \| \Mvisgt \odot (\Mxyzest - \Mxyzgt) \|_1
     + \| \Mvisest - \Mvisgt\|_1 \\ 
     +~& \lambda\| \Mfullest - \Mfullgt \|_1
     + CE(\Mvisgt \odot \Msraest, \Msragt ).
    \end{split}
\end{equation}
Thereby, $\odot$ denotes element-wise multiplication and we only supervise $M_\text{XYZ}$ and $M_\text{SRA}$ using the visible region. 
Specifically, for occluded datasets such as LM-O, we set $\lambda = 1$, while for occlusion-free datasets like LM, we set $\lambda = 0$.

The overall loss for GDRN can be summarized as $\loss_{\text{GDR}} = \loss_{\text{Pose}} + \loss_{\text{Geom}}.$
Notice that our GDRN can be trained end-to-end, without requiring any three-stage training strategy as in ~\cite{li2019cdpn}.

\noindent \textbf{Decoupling Detection and 6D Object Pose Estimation.}
Similar to~\cite{li2019cdpn,labbe2020CosyPose}, we mainly focus on the network for 6D object pose estimation and make use of an existing 2D object detector to obtain the zoomed-in input RoIs.
This allows us to directly make use of the advances in runtime~\cite{redmon2018yolov3,bochkovskiy2020yolov4,ge2021yolox} and accuracy~\cite{faster,tian2019fcos} within the rapidly growing field of 2D object detection, without having to change or re-train the pose network.
Therefore, we adopt a simplified Dynamic Zoom-In (DZI)~\cite{li2019cdpn} to
decouple the training of our GDRN and object detectors.
During training, we first uniformly shift the center and scale of the ground-truth bounding boxes by a ratio of 25\%. We then zoom in the input RoIs with a ratio of $r=1.5$ while maintaining the original aspect ratio. This ensures that the area containing the object is approximately half the RoI.
DZI can also circumvent the need of dealing with varying object sizes.

Noteworthy, although we employ a two-stage approach,
one could also implement GDRN on top of any object detector and train it in an end-to-end manner.

\subsection{Geometry-guided Pose Refinement}
\label{sec:refine}

\Rone{
To improve pose accuracy when depth information is available, we propose a novel pose refinement module.
Despite the advantages of the CIR~\cite{lipson2022coupled} mentioned in Sec.~\ref{rw_pose_refine}, it faces limitations when the rendered and observed images differ significantly due to variations in lighting conditions or object materials. 
These domain mismatches can impair the performance of the optical flow estimator.
To mitigate this issue, we incorporate the predicted coordinate map $\Mxyz$ from GDRN as an additional input to the optical flow estimator. 
Specifically, we utilize the coordinate map inferred from the input image and compare it with the coordinate map rendered based on the predicted pose to establish correspondences. 
This strategy provides domain-invariant information, improving robustness and mitigating the adverse effects of domain mismatches.
}

\Rone{
A straightforward approach to incorporate this information is directly concatenating the predicted coordinate map with images from other modalities as input. 
However, as demonstrated in our experiments, this method does not consistently lead to performance improvement. 
This limitation is primarily due to the possibility of inaccuracies in the predicted coordinate map, which can degrade overall performance.
To address this issue, we propose a more effective solution: instead of using the raw coordinate map directly, we extract features from the coordinate maps and assign a confidence weight to these features. 
The confidence weight is determined based on the discrepancy between the predicted and rendered coordinates maps. 
This approach enables the model to leverage the coordinate map effectively when it is accurate, while maintaining robustness in scenarios where the coordinate map contains errors.
}

\noindent \textbf{Problem formulation.} Given the observed image $\img_0=\img$, depth map $\depth_0$, and the outputs of GDRN, including 1) the pose prediction
$\pose_0^{(0)}$, 2) the predicted object coordinate map $\coor_0 = \Mxyz$, and 3) the predicted object masks $\mask_0=\Mvis$, the goal of the refinement module is to refine the pose iteratively, and to yield a final pose prediction $\pose_0^{(T)}$ after $T$ steps.

\noindent \textbf{Overview.}
\Rtwo{
Fig. \ref{fig:refinement_pipeline} presents a schematic overview of the proposed methodology.
}
In each iteration, given the initial pose $\pose_0^{(0)}$, we add perturbations on $\pose_0^{(0)}$ to generate a set of poses $\{\pose_i| i=1,2,..,n \}$ by adding or subtracting an angle $\theta$
from either roll, pitch, or yaw.
For each pose $\pose_i$, we render the image $\img_i$, depth map $\depth_i$, object mask $\mask_i$ and coordinate map $\coor_i$ of the object. 

\Rtwo{
In each iteration $t$, we refine the object pose by aligning the correspondences between $\img_0$ and $\{\img_i\}$ solved in two parallel ways following \cite{lipson2022coupled}.}
For each point $\point_i$ in the rendered image $\img_i$, we compute its corresponding point $\point_{i\rightarrow0}$ in the observed image $\img_0$ by (a) the previous object pose prediction $\pose_0^{(t)}$ or (b) predicted 3D optical flows.
Similarly, we compute the corresponding points $\point_{0\rightarrow i}$ in $\img_i$ of each point in $\img_0$.
We formulate the differences between (a) and (b) as the optimization objective and use the Gauss-Newton algorithm to optimize the pose prediction $\pose_0^{(t+1)}$ for the next iteration.
\Rtwo{
We repeat this optimization for $T = 10$ iterations (inner loop).  
Subsequently, a new set of poses $\{P_i\}$ is generated, and the corresponding new image set is rendered.  
This refinement process is repeated $N_\text{out}=4$ times (outer loop).
}

\noindent \textbf{Correspondences from the previous pose prediction.}
We designate the 3D coordinate of a point as $\point=[x,y,d]^\top$, where $x,y$ are the image coordinates normalized by $\mathbf{K}^{-1}$ and $d$ is the inverse depth value.
For a point $\point_{0}$ in the observed image $\img_0$, its corresponding point $\point_{0\rightarrow i}$ in the rendered image $\img_i$ can be computed by the previous pose prediction $\pose_0^{t}$ as
\begin{equation}
    \point_{0\rightarrow i}^{(\text{pose}, t)} = \Pi(\pose_i (\pose_0^{(t)})^{-1} \Pi^{-1}(\point_0)).
\end{equation}
where $t$ is the number of iterations, $\Pi$ and $\Pi^{-1}$ are the depth-augmented pinhole projection functions that convert coordinates of a point between the world frame $\mathbf{X}=[X,Y,Z]^\top$ and the normalized image frame $\point=[x,y,d]^\top$ as
\begin{equation}
\begin{split}
    \Pi(X) &= \frac{1}{Z}[X, Y, 1]^\top, \\
    \Pi^{-1}(\point) &= \frac{1}{d}[x, y, 1]^\top.
\end{split}
\end{equation}
Analogously, for $\point_i$ in the rendered image $\img_i$, its corresponding point in $\img_0$ is 
\begin{equation}
    \point_{i\rightarrow0}^{(\text{pose},t)} = \Pi(\pose_0^{(t)} (\pose_i)^{-1} \Pi^{-1}(\point_i)).
\end{equation}

\begin{figure}[t]
    \begin{center}
    \includegraphics[width=0.96\linewidth]{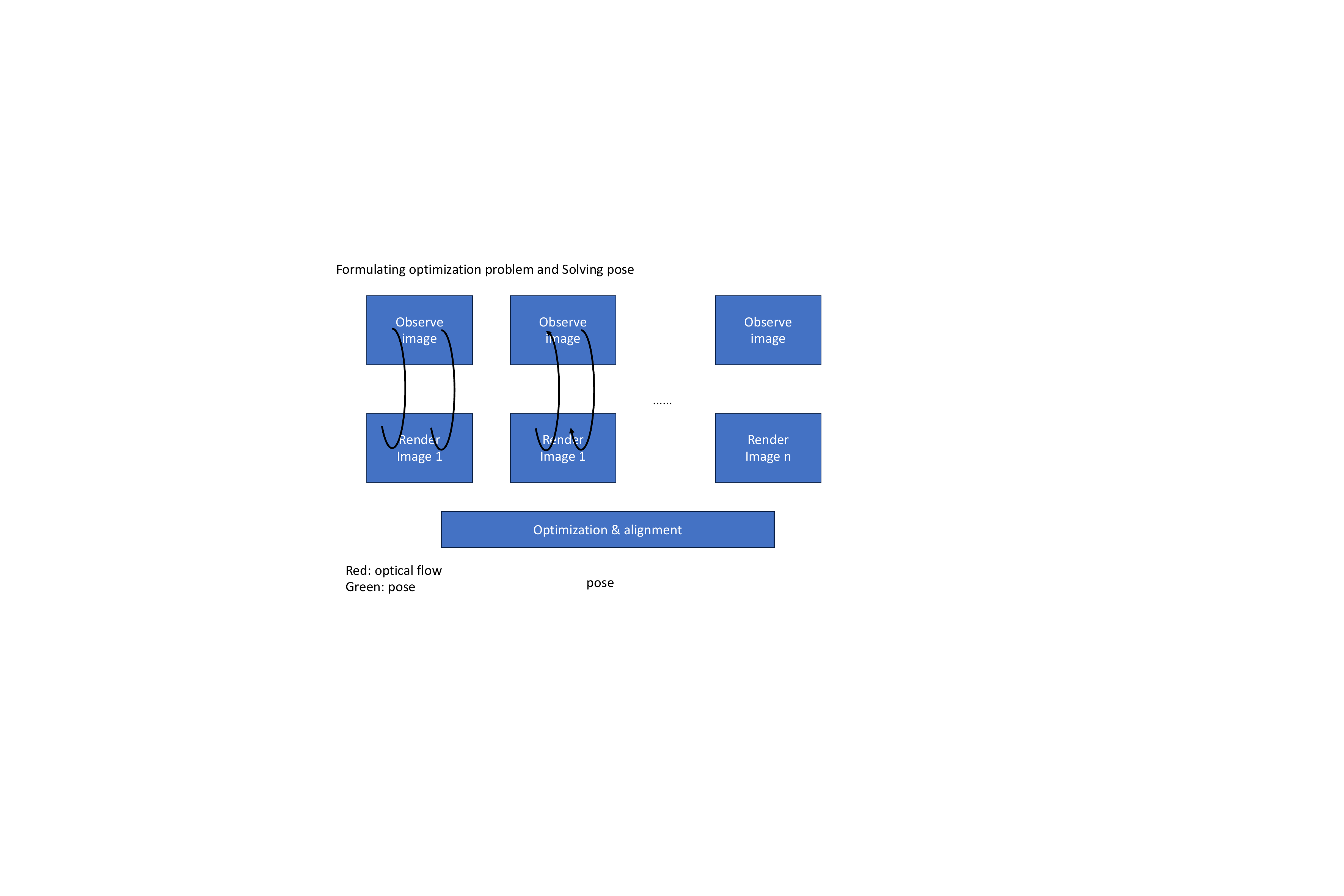}
    \end{center}
    \caption{
        \label{fig:optical_flow}
        \textbf{Overview of the 3D optical flow estimator.}
        We first use the correspondences inferred from the previous pose prediction to sample the rendered coordinate map $\mathbf{C}_i$ and get $\mathbf{C}'_0$.
        Then we concatenate the predicted coordinate map $\mathbf{C}_0$ and $\mathbf{C}_0'$ and mask the visible region.
        The coordinate feature $\mathbf{c}_{0 \rightarrow i}$ is extracted by a convolutional network $\Lambda$ and weighted dynamically by $\mathbf{\omega}_c$ according to the quality of the coordinate map.
        The weighted coordinate feature, context feature, depth feature, the correlation feature $\mathbf{s}_{0 \rightarrow i}$, along with the hidden state $\mathbf{h}_{0 \rightarrow i}$ are fed into the GRU-based update module, which outputs the correspondences $\mathbf{x}_{0 \rightarrow i}^{(\text{flow}, t)}$ and a new hidden state $\mathbf{h}_{0 \rightarrow i}^{(t)}$.
        The correspondences $\mathbf{x}_{i \rightarrow 0}^{(\text{flow}, t)}$ are calculated in a symmetric manner.
    }
\end{figure}

\noindent \textbf{Correspondences from optical flows.}
Our goal is to refine the previous pose prediction by establishing accurate and robust 3D-3D correspondences.
To this end, we predict 3D optical flow with the guidance of the coordinate map.
We define the 3D optical flow as $\Delta \point=[\Delta x, \Delta y, \Delta d]^\top$ in this paper, which consists of the traditional 2D optical flow and the motion of the inverse depth.
The overview of the 3D optical flow estimator is shown in Fig.~\ref{fig:optical_flow}.
We use $\point_{0\rightarrow i}^{(\text{pose},t)}$ as the initialization of $\point_{0\rightarrow i}^{(\text{flow},t)}$, and refine the correspondences by predicting 3D optical flow residuals $\mathbf{r}_{0\rightarrow i}^{(t)}$ and $\mathbf{r}_{i\rightarrow0}^{(t)}$, denoted as
\begin{equation}
\begin{split}
    \point_{0\rightarrow i}^{(\text{flow}, t)} = \point_{0\rightarrow i}^{(\text{pose}, t)} + \mathbf{r}_{0\rightarrow i}^{(t)}, \\
    \point_{i\rightarrow0}^{(\text{flow}, t)} = \point_{i\rightarrow0}^{(\text{pose}, t)} + \mathbf{r}_{i\rightarrow0}^{(t)}.
\end{split}
\end{equation}

\noindent \Rtwo{\textbf{3D optical flow estimator.}}
\Rtwo{
We provide an overview of the optical flow estimator in Fig. \ref{fig:optical_flow}.}
Built upon RAFT~\cite{teed2020raft}, we propose a coordinate-augmented RAFT to predict the optical flow residuals $\mathbf{r}_{0\rightarrow i}^{(t)}, \mathbf{r}_{i\rightarrow0}^{(t)}$ along with their confidence weight maps.

Following CIR \cite{lipson2022coupled}, a GRU-based update module is employed for iterative optical flow estimation. 
\Rtwo{
Given an image-render pair $\{\img_0, \img_i\}$, we first extract features from $\img_0$ and $\img_i$ using a convolutional network and set up a correlation pyramid, as in RAFT.
Using the lookup operator, we retrieve the correlation feature $\mathbf{s}_{0\rightarrow i}$ where $\point_{0\rightarrow i}^{(\text{pose}, t)}$ serves as the index.
}

During each iteration, the update module is fed with the correlation feature $\mathbf{s}_{0\rightarrow i}$, the weighted coordinate feature $\omega_c \mathbf{c}_{0\rightarrow i}$ \Rtwo{(introduced below)}, the previous hidden state $\mathbf{h}_{0\rightarrow i}^{(t-1)}$, and the context and depth features.
The initial hidden state $\mathbf{h}_{0\rightarrow i}^{(0)}$ as well as the context and depth features are computed in accordance with CIR.
The update module outputs a new hidden state $\mathbf{h}_{0\rightarrow i}^{(t)}$, the optical flow residuals $\mathbf{r}_{0\rightarrow i}^{(t)}$, and a dense confidence map $\mathbf{w}_{0\rightarrow i}^{(t)}$. 
The confidence map dynamically identifies outliers, improving the robustness of correspondences.
Similarly, the same update module is applied in the reverse direction, using $\mathbf{s}_{i\rightarrow 0}$ and $\mathbf{c}_{i\rightarrow 0}$ to predict $\mathbf{r}_{i\rightarrow 0}^{(t)}$ and $\mathbf{w}_{i\rightarrow 0}^{(t)}$.

\noindent \Rtwo{
\textbf{Leveraging coordinate map for optical flow estimation.}
We encode the predicted coordinates from GDRN into coordinate features to provide domain-invariant information.
Since $C_0$ and $C_i$ belong to the same domain, it is unnecessary to set up a correlation volume or use a lookup operator to retrieve features as done for RGB images.
Instead, we bilinearly sample $\coor_i$ using $\point_{0\rightarrow i}^{(\text{pose}, t)}$ as the index to generate a coordinate map $\coor'_0$.
This map transforms $\coor_i$ into $\coor_0$ based on the current predicted optical flow.
By comparing $\coor'_0$ with $\coor_0$, we can assess the accuracy of the optical flow prediction and further refine it.}

\Rtwo{
However, since some points in $\img_0$ are not visible in $\img_i$ and therefore lack valid correspondences, it is necessary to isolate the object regions visible under both poses and mask out outliers. 
To achieve this,} we sample the mask $\mask_i$ using $\point_{0 \rightarrow i}^{(\text{pose}, t)}$ to generate a corresponding mask $\mask'_0$.
To guide the prediction of optical flow residuals, we use a convolutional network $\Lambda$ to encode the difference of $\coor'_0$ and $\coor_0$ into a coordinate feature $\mathbf{c}_{0\rightarrow i}$.
The coordinate feature $\mathbf{c}_{0\rightarrow i}$ is computed as follows,
\begin{equation}
    \mathbf{c}_{0\rightarrow i} =  \Lambda( \mathbf{M}'_0 \odot \mathbf{M}_0 \odot (\coor'_0~ \text{\scriptsize\sffamily\textcopyright}~ \coor_{0})),
    \label{eq:coor_feat}
\end{equation}
where {\scriptsize \sffamily\textcopyright} is the concatenation operator and $\odot$ is the element-wise production.
\Rtwo{
We use $\mask_0$ and $\mask'_0$ to ensure that only the object regions visible under both poses contribute to the feature computation, while outliers are effectively masked out.}

The quality of the coordinate $\mathbf{C}_0$ predicted by GDRN significantly influences the quality of $\mathbf{c}_{0\rightarrow i}$ and the precision of the pose prediction.
\Rtwo{
Inaccuracies in the predicted coordinate map can degrade overall performance.
To ensure robustness, we introduce a confidence weight for the coordinate feature $\mathbf{c}_{0\rightarrow i}$.}
The confidence weight $\omega_c$ is defined as 
\begin{equation}
    \omega_c = \mathbbm{1}(\text{avg}(\mathbf{M}'_0 \odot \mathbf{M}_0 \odot |\mathbf{C}'_0 - \mathbf{C}_0|) < \gamma),
\label{eq:weight}
\end{equation}
where $\mathbbm{1}(\bullet)$ is the indicator function, $\text{avg}(\bullet)$ computes the average error between $\mathbf{C}'_0$ and $\mathbf{C}_0$, and $\gamma$ is a threshold hyperparameter.
\Rtwo{
If the average error exceeds the threshold $\gamma$, the coordinate is deemed unreliable and $\omega_c$ is set to 0.
Otherwise, $\omega_c=1$.
The weighted coordinate feature is then computed as $\omega_c \mathbf{c}_{0\rightarrow i}$, ensuring that only reliable coordinate features contribute to the optical flow refinement process, thereby enhancing the robustness of the overall system.
}

\noindent \textbf{Optimization.}
The optimization objective is defined as follows
\begin{equation}
\begin{split}
    \mathop{\arg \min}\limits_{\pose_0^{(t)} \in SE(3)} \mathcal{E}(\pose_0^{(t)})=
    \sum_{i=1}^{n} \sum_{\point_0 \in \mask_0} \mathbf{w}_{0\rightarrow i}^{(t)} \|\point_{0\rightarrow i}^{(\text{flow},t)} - \point_{0\rightarrow i}^{(\text{pose},t)} \|^2 \\
    + \sum_{i=1}^{n} \sum_{\point_i \in \mask_i} \mathbf{w}_{i\rightarrow0}^{(t)} \|\point_{i\rightarrow0}^{(\text{flow},t)} - \point_{i\rightarrow0}^{(\text{pose},t)} \|^2  ,
\end{split}
\label{eq:goal}
\end{equation}
where $\| \bullet \|$ is the Euclidean distance and $\mask_i$ is the object mask of $\img_i$.

The objective defined in Eq.~\ref{eq:goal} aims to find camera poses $\pose_0$ that result in reprojected points $\point_{i\rightarrow0}^{(pose)}, \point_{0\rightarrow i}^{(pose)}$ that align with the revised correspondences $\point_{i\rightarrow0}^{(flow)}, \point_{0\rightarrow i}^{(flow)}$.
We compute the gradient of $\pose_0^{(t)}$ in $ \point_{0\rightarrow i}^{(\text{pose},t)}, \point_{i\rightarrow 0}^{(\text{pose},t)}$ and perform three steps of Gauss-Newton updates to obtain $\pose_0^{(t+1)}$.

\noindent \textbf{Training.}
For supervision, we evaluate the predicted optical flow and refined pose estimates from all update iterations in the forward pass. 
Specifically, we use $\mathcal{L}_\text{Pose}$ in Eq.~\ref{eq:loss_pose} to supervise the estimated pose and employ the $\mathcal{L}_1$ endpoint error as the loss to supervise the optical flow.
During training, we introduce random perturbations to the ground-truth rotation and translation for pose initialization. 
We generate the input coordinate by first rendering the coordinate map with a perturbed pose and then adding Gaussian noise.
We perform one outer iteration and render only one image for $\pose_0^{(0)}$ at each training step.

\noindent \textbf{Handling symmetry.}
For symmetric objects, the coordinate map rendered by the predicted pose might be inconsistent with the predicted coordinate map.
Therefore, given the set of all possible poses under symmetry $\mathcal{P}$, we select the pose with the most similar rendered coordinate map before feeding it to the refinement module.
Concretely, the selected pose is
\begin{equation}
    \pose_0^{(0)} = \underset{\pose \in \mathcal{P}}{\arg\min}(\text{avg}|\Theta(\pose) - \coor_0|),
    \label{eq:sym_pose}
\end{equation}
where $\Theta$ is the rendering function to get the rendered coordinate map given an object pose.


\begin{figure}
    \centering
    \begin{subfigure}[b]{0.4\linewidth}
        \centering
        \includegraphics[width=\linewidth]{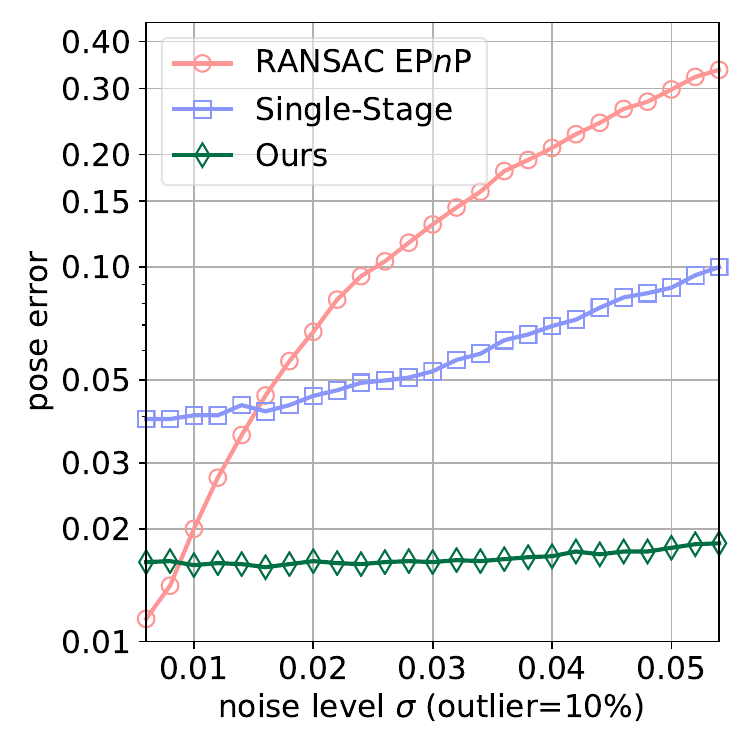}\vspace{-2mm}
        \caption{\label{fig:outlier_10}}
    \end{subfigure}
    \begin{subfigure}[b]{0.4\linewidth}
        \centering
        \includegraphics[width=\linewidth]{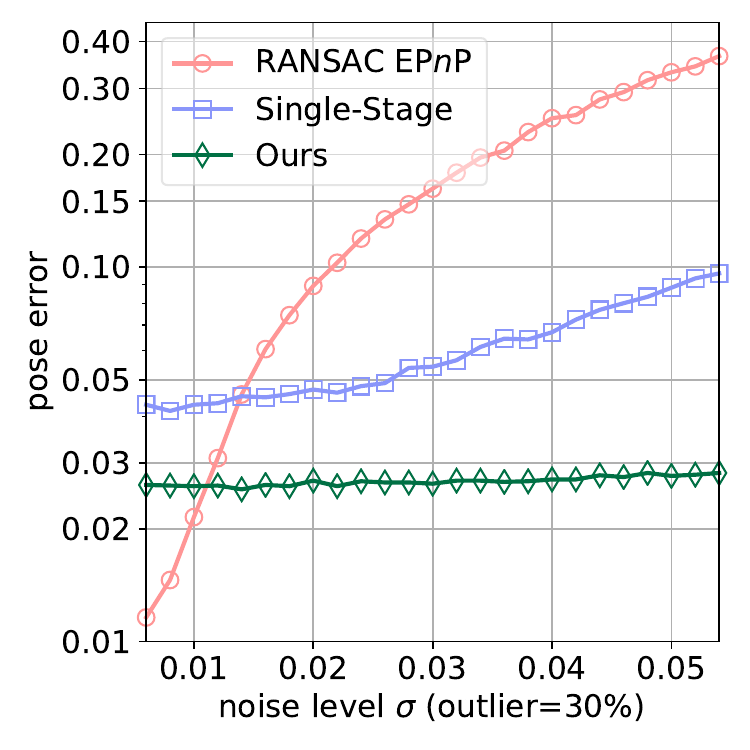}\vspace{-2mm}
        \caption{\label{fig:outlier_30}}
    \end{subfigure}
    \begin{subfigure}[b]{0.18\linewidth}
        \centering
        \includegraphics[width=\linewidth]{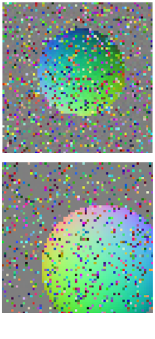}\vspace{-2mm}
        \caption{\label{fig:sphere_input}}
    \end{subfigure}\vspace{-2mm}
\caption{
\label{fig:sphere_outlier_plot}%
            {\bf Results of P$n$P variants on Synthetic Sphere.}
            \emph{(a, b)}:
            We compare our Patch-P$n$P module with the traditional RANSAC EP$n$P~\cite{lepetit2009epnp} and
            another learning-based P$n$P~\cite{hu2020single_stage}.
            The pose error is reported as relative ADD error \wrt the sphere's diameter (y-axis in log-scale).
            \emph{(c)}: Zoomed-In ($64\times64$) synthetic examples for Patch-P$n$P.
}
\end{figure}

\section{Experiments}

In this section, we first introduce our experimental setup and then present the evaluation results for several commonly employed benchmark datasets.
Thereby, we first present experiments on a synthetic toy dataset, which clearly demonstrates the benefit of our Patch-P$n$P compared to the classic optimization-driven P$n$P.
Additionally, we demonstrate the effectiveness of our individual components by performing ablative studies on LM~\cite{Hinterstoisser2012} and LM-O~\cite{Brachmann2014Learning6O}.
Finally, we compare our method with state-of-the-art methods on the BOP benchmark~\cite{sundermeyer2023bop}, which contains seven core datasets including LM-O~\cite{Brachmann2014Learning6O}, YCB-V~\cite{xiang2017posecnn}, T-LESS~\cite{hodan2017t}, TUD-L~\cite{hodan2018bop}, IC-BIN~\cite{doumanoglou2016recovering}, ITODD~\cite{drost2017introducing} and HB~\cite{kaskman2019homebreweddb}.

\subsection{Experimental Setup}
\noindent \textbf{Implementation Details.}
All our experiments are implemented using PyTorch~\cite{paszke2019pytorch}.
We train the GDRN(PP) end-to-end using the Ranger optimizer~\cite{liu2019radam,zhang2019lookahead,GradientCentra} which combines the RAdam~\cite{liu2019radam} optimizer with Lookahead~\cite{zhang2019lookahead} and Gradient Centralization~\cite{GradientCentra} on a single NVIDIA 3090 GPU.
On the LM dataset, we set the total training epoch to 160 with a batch size of 24 and a base learning rate of $10^{-4}$, which we anneal at 72\,\% of the training phase using a cosine schedule~\cite{loshchilov-ICLR17SGDR}.
While for the BOP datasets, we train GDRN for 40 epochs under the one model per dataset setting, and 100 epochs under the one model per object setting, with a batch size of 36 and a base learning rate of $8\times10^{-4}$.
The refinement module is trained from scratch using the AdamW~\cite{adamw} optimizer for 200k steps with batch size 12 for each dataset on 2 NVIDIA 3090 GPUs. 
We adopt an exponential learning rate schedule with a linear increase to $3 \times 10^{-4}$ over the first 10k steps and a 50\,\% drop for every 20k steps afterwards, 
and the weight decay is set to $10^{-5}$.

\noindent \textbf{Datasets.}
We conduct our experiments on nine datasets:
Synthetic Sphere~\cite{lepetit2009epnp,hu2020single_stage},
LM~\cite{Hinterstoisser2012}
and seven core datasets included in the BOP benchmark~\cite{hodan_bop20}.
\emph{The Synthetic Sphere dataset} contains 20k samples for training and 2k for testing,
created by randomly capturing a unit sphere model
using a virtual calibrated camera with a focal length of 800, resolution 640$\times$480,
and the principal point located at the image center.
The Rotations and translations are uniformly sampled in 3D space, and within an interval of $[-2,2]\times[-2,2]\times[4,8]$, respectively.
\emph{LM dataset} consists of 13 sequences,
each containing $\approx$~1.2k images with ground-truth poses for a single object with clutter and mild occlusion.
We follow~\cite{brachmann2016uncertainty} and employ $\approx$15\,\% of the RGB images for training and 85\,\% for testing.
We additionally use 1k rendered RGB images for each object during training as in~\cite{li2019cdpn}.
\emph{LM-O} consists of 1214 images from an LM sequence,
where the ground-truth poses of 8 visible objects with more occlusion are provided for testing.
\emph{YCB-V} is a very challenging dataset exhibiting strong occlusion, clutter and several symmetric objects.
It comprises over 110k real images captured with 21 objects, both with and without texture.
\emph{T-LESS} contains 30 industry-relevant objects that lack significant texture or discriminative color. It is quite challenging due to object symmetries and mutual similarities between objects.
\emph{TUD-L} comprises three moving objects captured under diverse lighting conditions and varying degrees of occlusion.
\emph{IC-BIN} provides a comprehensive collection of cluttered scenes involving two objects with heavy occlusion, specifically designed for evaluating pose estimation in the bin-picking scenario.
\emph{ITODD} comprises grayscale images captured in realistic industrial scenarios, featuring a diverse collection of 28 textureless objects.
\emph{HB} consists of 33 objects captured in 13 scenes, each exhibiting varying levels of complexity.
For all the seven BOP core datasets, we also leverage the publicly available synthetic data using physically-based rendering (\texttt{pbr})~\cite{hodan_bop20} for training.

\begin{table*}[t]
    \centering
    \caption{
        {\bf Ablation study on LM.}
        \Rthree{\emph{(a):} Ablation of number of regions in $\Msra$.}
        \emph{(b):} Ablation of P$n$P type, the parameterization of $\rot$ and $\trans$, loss type and
            geometric guidance.
    }
    \begin{subfigure}[t]{0.25\linewidth}
        \vspace{0pt}
        \includegraphics[width=\linewidth]{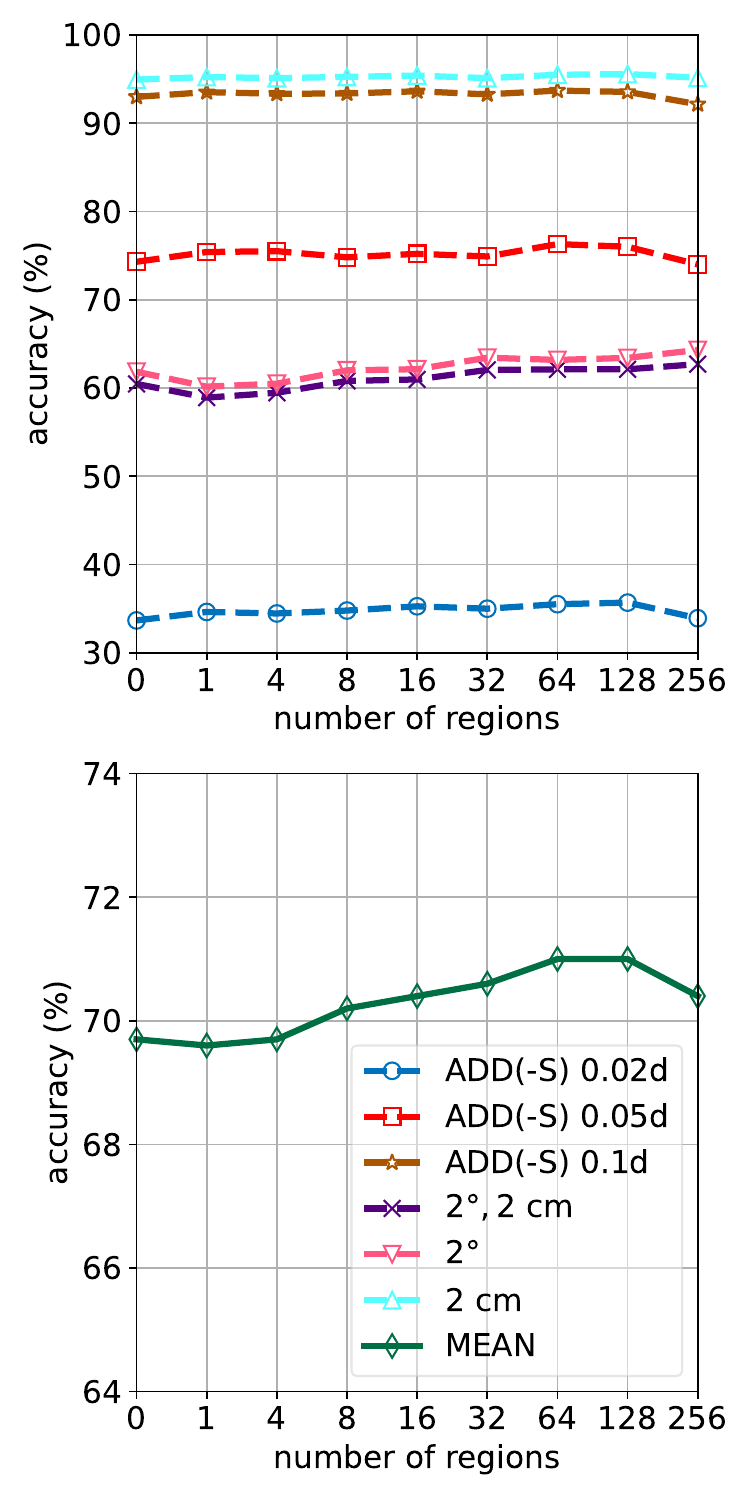}\vspace{-3.5mm}
        \caption{\label{fig:ablate_region}}
    \end{subfigure}~~
    \begin{subfigure}[t]{0.75\linewidth}
    \vspace{3pt}
    \tablestyle{2.5pt}{1.1}
	\begin{tabular}{l | l | ccc | c | c | c | c}
    \multirow{2}{*}{Row} & \multirow{2}{*}{Method} & \multicolumn{3}{c|}{ADD(-S)}& \multirow{2}{*}{$2\degree, 2~\text{cm}$} & \multirow{2}{*}{$2\degree$} & \multirow{2}{*}{$2~\text{cm}$} & \multirow{2}{*}{MEAN} \\
                         & & 0.02d&0.05d & 0.1d &  & & &  \\
	\shline
    A0& CDPN~\cite{li2019cdpn} & - &-    & 89.9 & -    & -    & 92.8 & -      \\
   \hline
    B0& GDRN (\textbf{Ours})                                              & 35.5 & 76.3 & 93.7 & 62.1 & 63.2 & 95.5 & \textbf{71.0}   \\
    B1& B0: $\rightarrow$ Test with P$n$P/RANSAC                          & 31.0 & 72.1 & 92.2 & 67.1 & 68.9 & 94.5 & \textbf{71.0}  \\
    B2& B0: $\rightarrow$ Patch-P$n$P for $\trans$; P$n$P/RANSAC for $\rot$& 35.6& 76.0 & 93.6 & 67.1 & 69.0 & 95.5 & \textbf{72.8}  \\
   \hline
    C0& B0: Patch-P$n$P $\rightarrow$ PointNet-like PnP                   & 29.2 & 72.6 & 92.3 & 44.5 & 45.8 & 94.3 & 63.1  \\
    C1 & B0: Patch-P$n$P $\rightarrow$ BP$n$P~\cite{chen2020BPnP}         & 34.3 & 72.6 & 92.0 & 64.3 & 66.0 & 94.4 & 70.6 \\
    \hline
    D0& B0: \Rthree{Allocentric $\rot_\text{6d}$} $\rightarrow$ Egocentric $\rot_\text{6d}$   & 36.1 & 75.7 & 93.2 & 60.4 & 61.5 & 95.3 & 70.4  \\
    D1& B0: \Rthree{Allocentric $\rot_\text{6d}$} $\rightarrow$ Allocentric quaternion        & 24.8 & 67.4 & 90.5 & 35.5 & 36.9 & 92.2 & 57.9  \\
    D2& B0: \Rthree{Allocentric $\rot_\text{6d}$} $\rightarrow$ Allocentric log quaternion    & 22.7 & 64.6 & 88.9 & 33.7 & 35.4 & 90.9 & 56.0  \\
    D3& B0: \Rthree{Allocentric $\rot_\text{6d}$} $\rightarrow$ Allocentric Lie algebra vector& 23.0 & 66.3 & 89.7 & 33.8 & 35.3 & 91.4 & 56.6  \\
    \hline
    E0& B0: $\trans_\text{SITE}$ $\rightarrow$ $\trans$                   & 28.3 & 72.0 & 92.4 & 61.6 & 63.2 & 94.6 & 68.7  \\
    E1& B0: $\trans_\text{SITE}$ $\rightarrow$ $(o_x, o_y); t_z$          & 31.4 & 73.7	& 93.3 & 50.4 & 51.6 & 94.7 & 65.8  \\
    E2& B0: $\delta_z$ $\rightarrow$ $t_z$                                & 32.8 & 73.5 & 93.3 & 63.3 & 64.8 & 94.9 & 70.4  \\
    \hline
    F0& B0: $\loss_\text{Pose}$ $\rightarrow$ {\fontsize{7pt}{0pt}\selectfont $\loss_\text{PM} = \underset{\mathbf{x} \in \mathcal{M}}{\avg} \| (\hat{\rot} \mathbf{x} + \hat{\trans}) - (\bar{\rot} \mathbf{x} + \bar{\trans}) \|_1$}
                                                                          & 33.7 & 76.5 & 94.1 & 47.4 & 48.2 & 95.8 & 65.9  \\
    F1& F0: $\loss_\text{PM}$ $\rightarrow$ Disentangling $\rot; \trans$  & 30.8 & 71.1 & 91.8 & 64.6 & 66.8 & 93.5 & 69.8  \\
    F2& F0: $\loss_\text{PM}$ $\rightarrow$ Disentangling $\rot; (t_x, t_y); t_z$
                                                                          & 32.2 & 73.9 & 93.6 & 63.8 & 65.3 & 94.8 & 70.6  \\

    F3& B0: $\loss_\rot$ $\rightarrow$ Angular loss                       & 32.4 & 75.5 & 93.8 & 40.2 & 40.9 & 95.7 & 63.1  \\
    F4& B0: $\loss_\rot$ $\rightarrow$ $\loss_{\rot,\text{sym}}$          & 35.5 & 75.8 & 93.9 & 61.6 & 62.7 & 95.4 & 70.8  \\
    \hline
    G0& B0: $\loss_\text{GDR}$ $\rightarrow$ w/o $\loss_\text{Geom}$      & 30.8 & 72.7 & 92.2 & 45.9 & 46.8 & 94.1 & 63.7  \\
    G1& G0: $\rightarrow$ w/o $\mathbf{M}_\text{2D}$                              & 18.6 & 60.1 & 85.6 & 26.0 & 27.8 & 87.6 & 51.0  \\
    G2& G0: $\rot_\text{a6d}$ $\rightarrow$ Allocentric quaternion        & 6.7  & 40.6 & 73.2 & 6.2  & 7.4  & 75.6 & 34.9  \\
    \hline
    H0& B0: Faster R-CNN~\cite{faster} $\rightarrow$ YOLOv3~\cite{redmon2018yolov3} & 33.9 & 75.6 & 93.7 & 60.9&62.1&95.2&70.2   \\
    \end{tabular}
\caption{\label{tab:lm_ablate}}
\end{subfigure}\hfill
\end{table*}

\noindent \textbf{Evaluation Metrics.}
We use three common metrics for 6D object pose evaluation, \ie~ADD(-S)~\cite{Hinterstoisser2012,hodan2016evaluation}, 
$n\degree, n\,\mathrm{cm}$~\cite{Shotton_2013_CVPR_NcmNdegree} and the BOP metric~\cite{hodan2018bop,hodan_bop20,sundermeyer2023bop}.
The \emph{ADD} metric measures whether the average deviation of the transformed model points is
less than 10\,\% of the object's diameter (0.1d).
For symmetric objects, the \emph{ADD-S} metric is employed to measure the error as the average distance to the closest model
point~\cite{Hinterstoisser2012,hodan2016evaluation}.
The \emph{$n\degree, n\,\mathrm{cm}$} metric
measures whether the rotation error is less than $n\degree$ and the translation error is below $n\,\mathrm{cm}$.
Notice that to account for symmetries, $n\degree, n\,\mathrm{cm}$ is computed \wrt the smallest
error for all possible ground-truth poses~\cite{li2019deepim}.
The \emph{BOP} metric is a symmetry-aware comprehensive metric, which is calculated as the mean of the Average Recall of three metrics: $\text{AR}_\text{{\tiny BOP}}=(\text{AR}_\text{\tiny MSPD} + \text{AR}_\text{\tiny MSSD} + \text{AR}_\text{\tiny VSD})/3$.
Please refer to~\cite{hodan_bop20} for a detailed explanation of these metrics.


\subsection{Toy Experiment on Synthetic Sphere}
We conduct a toy experiment comparing our approach
with P$n$P/RANSAC and~\cite{hu2020single_stage} on the Synthetic Sphere dataset.
We generate $\Mxyz$ from the provided poses and feed them to our Patch-P$n$P.
For fairness, $\Msra$ is excluded from the input.
Following~\cite{hu2020single_stage},
during training, we randomly add Gaussian noise $\mathcal{N}(0, \sigma^2)$ with $\sigma\in\mathcal{U}[0, 0.03]$ to each point of the dense coordinates maps.
Since the coordinates maps are normalized in $[0, 1]$,
we choose 0.03 as it reflects approximately the same level of noise as in~\cite{hu2020single_stage}.
Additionally, we randomly generated 0\,\% to 30\,\% of outliers for $\Mxyz$ (Fig.~\ref{fig:sphere_input}).
During testing, we report the relative ADD error \wrt the sphere's diameter on the test set with different levels of noise and outliers.

\noindent \textbf{Comparison with P$n$P/RANSAC and \cite{hu2020single_stage}.}
In Fig.~\ref{fig:sphere_outlier_plot}, we demonstrate the effectiveness and robustness of our approach by comparing Patch-P$n$P with the traditional RANSAC-based EP$n$P~\cite{lepetit2009epnp} and
the learning-based P$n$P from~\cite{hu2020single_stage}).
As depicted in Fig.~\ref{fig:sphere_outlier_plot},
while RANSAC-based EP$n$P\footnote{We follow the state-of-the-art method CDPN~\cite{li2019cdpn} for the implementation and hyper-parameters of P$n$P/RANSAC in all our experiments.} is more accurate when noise is unrealistically minimal, learning-based P$n$P methods are much more accurate and robust as the level of noise increases.
Moreover, Patch-P$n$P is significantly more robust than Single-Stage~\cite{hu2020single_stage} \wrt to noise and outliers, thanks to our geometrically rich and dense correspondences maps.


\subsection{Ablation Study on LM}
\label{sec:ablation}
We present several ablation experiments for the widely used LM dataset~\cite{Hinterstoisser2012}.
We train a single model for all objects for 160 epochs without applying any color augmentation.
For fairness in evaluation, we leverage the detection results from Faster R-CNN as provided by~\cite{li2019cdpn}.

\noindent \textbf{Number of Regions in $\Msra$.}
In Table~\ref{fig:ablate_region}, we show results for different numbers of regions in $\Msra$.
Thereby, without our attention $\Msra$ (number of regions = 0), the accuracy is deliberately good, which suggests the effectiveness and versatility of Patch-P$n$P. 
Nevertheless, the overall accuracy can be further improved with increasing number of regions in $\Msra$, despite starting to saturate around 64 regions.
Thus, we use 64 regions for $\Msra$ in all other experiments as a trade-off between accuracy and memory.

\noindent \textbf{Effectiveness of Patch-P$n$P.}
We demonstrate the effectiveness of the image-like geometric features ($\Mcorr, \Msra$) by comparing
our Patch-P$n$P with traditional P$n$P/RANSAC~\cite{li2019cdpn}, the PointNet-like~\cite{qi2017pointnet} P$n$P from~\cite{hu2020single_stage}, and a differentiable P$n$P (BP$n$P~\cite{chen2020BPnP}).
For PointNet-like P$n$P, we extend the PointNet in~\cite{hu2020single_stage} to account for dense correspondences.
Specifically, we utilize PointNet to pointwisely transform the spatially flattened geometric features ($\Mcorr$ and $\Msra$) and directly predict the 6D pose
with global max pooling followed by two FC layers.
Since the correspondences are explicitly encoded in $\Mcorr$, no special attention is needed for the keypoint orders as in~\cite{hu2020single_stage}.
For BP$n$P~\cite{chen2020BPnP}, we replace the Patch-P$n$P in our framework with their implementation of BP$n$P\footnote{\href{https://github.com/BoChenYS/BPnP}{https://github.com/BoChenYS/BPnP}}. 
As BP$n$P was originally designed for sparse keypoints, we further adapt it appropriately to deal with dense coordinates.

As shown in Table~\ref{tab:lm_ablate}, Patch-P$n$P is more accurate than traditional P$n$P/RANSAC (B0 \vs A0),
PointNet-like P$n$P (B0 \vs C0) and BP$n$P (B0 \vs C1) in estimating the 6D pose.
Furthermore, in terms of rotation, our Patch-P$n$P outperforms PointNet-like P$n$P by a large margin,
which proves the importance of exploiting the ordering within the correspondences. 
Noteworthy, Patch-P$n$P is much faster in inference and up to 4$\x$ faster in training than BP$n$P, since the latter relies on P$n$P/RANSAC for both phases.

\noindent \textbf{Parameterization of 6D Pose.}
In Table~\ref{tab:lm_ablate}, we illustrate the impact of our proposed 6D pose parameterization.
In particular, the 6-dimensional $\rot_{\text{6d}}$ (Eq.~\ref{eq:r6})
achieves a much more accurate estimate of $\rot$ than commonly used representations such as
unit quaternions~\cite{xiang2017posecnn,li2019deepim},
log quaternions~\cite{park2020latentfusion} and the Lie algebra-based vectors~\cite{lienet_do2018}  (\textit{c.f.} B0 \vs D1-D3, and G0 \vs G2).
Moreover, we can deduce that the allocentric representation is significantly stronger than the egocentric formulation (B0 \vs D0). %

Similarly, the parameterization of the 3D translation is of high importance. 
Essentially, directly predicting $\trans$ in 3D space leads to worse results than leveraging the scale-invariant formulation $\trans_\text{SITE}$ (E0 \vs B0).
Additionally, replacing the scale-invariant $\delta_z$ in $\trans_\text{SITE}$ with the absolute distance $t_z$
or directly regressing the object center $(o_x, o_z)$ leads to inferior poses \wrt translation (B0 \vs E1, E2).
Hence, when dealing with zoomed-in RoIs, it is essential to parameterize the 3D translation in a scale-invariant fashion.

\noindent \textbf{Ablation on Pose Loss.}
As mentioned in Section~\ref{sec:revisit}, the loss function has an impact on direct 6D pose regression.
In TABLE~\ref{tab:lm_ablate}, we compare our disentangled $\loss_{\text{Pose}}$ to a simple angular loss and
the Point-Matching loss~\cite{li2019deepim} (F0).
Furthermore, we present its disentangled versions following~\cite{Simonelli2019_disentangle_loss}.
As shown in (B0 and F0-F4), all variants of the PM loss are clearly better than the angular loss in terms of rotation estimation. 
In addition, disentangling the rotation $\rot$ and distance $t_z$ in $\loss_\text{PM}$ largely enhances the rotation accuracy. 
Nonetheless, the overall performance is slightly inferior to our disentangled formulation $\loss_\text{Pose}$, which disentangles $\trans_\text{SITE}$ rather than the 3D translation $\trans$.
It is worth noting that $\loss_{\rot,\text{sym}}$ has a rather insignificant contribution compared with $\loss_\rot$. 
This can be accounted to the lack of severe symmetries in LM and to our proposed surface region attention $\Msra$.

\noindent \textbf{Effectiveness of Geometry-Guided Direct Regression.}
Furthermore, we train GDRN leveraging only our pose loss $\loss_\text{Pose}$ by discarding the geometric supervision $\loss_\text{Geom}$.
Surprisingly, even the simple version outperforms CDPN~\cite{li2019cdpn}  \wrt ADD(-S) 0.1d, when employing $\rot_\text{6d}$ for rotation (TABLE~\ref{tab:lm_ablate} G0 \vs A0).
Yet, we clearly outperform our baseline using GDRN with explicit geometric guidance.
If we predict the rotation as allocentric quaternions, the accuracy decreases (G2 \vs G0),
which can partially account for the weak performance of
previous direct methods~\cite{xiang2017posecnn,lienet_do2018}.
Moreover, when we remove the guidance of $\mathbf{M}_\text{2D}$, the accuracy drops significantly (G0 \vs G1).
Based on these results, we can see that appropriate geometric guidance is essential for direct 6D pose regression.

Direct pose regression also enhances the learning of geometric features as the error signal from the pose can be backpropagated.
TABLE~\ref{tab:lm_ablate} (B1, B2) shows that when evaluating GDRN with
P$n$P/RANSAC from the predicted $\Mcorr$, the overall performance exceeds CDPN~\cite{li2019cdpn}.
Similar to CDPN, we run tests using P$n$P/RANSAC for $\rot$ and Patch-P$n$P for $\trans$,
which achieves the overall best accuracy (B2).
This demonstrates that our unified GDRN can leverage the best of both worlds, namely, geometry-based indirect methods and direct methods.

\noindent \textbf{Effectiveness of Detection and Pose Decoupling.}
Similar to CDPN~\cite{li2019cdpn}, we decouple the detector and GDRN by means of Dynamic Zoom-In (DZI).
When evaluating GDRN with the YOLOv3 detections from~\cite{li2019cdpn}, the overall accuracy only drops slightly while the accuracy for ADD(-S) 0.1d almost remains unchanged (TABLE~\ref{tab:lm_ablate} H0).

\begin{table}
\centering
\caption{
\label{tab:lmo_abla}
{\bf Ablation study on LM-O for GDRN.}
We report the Average Recall (\%) of the BOP metric.
Note that only synthetic data is used for training.
}
\tablestyle{3.4pt}{1.2}
\begin{tabular}{l |l | c c c | c}
Row & Method & 
$\text{MSPD}$ &
$\text{MSSD}$ &
$\text{VSD}$ &
$\text{AR}_\text{{\tiny BOP}}$  \\
\shline
A0 & baseline &  80.8 & 55.6 & 43.6 & 60.0 \\
\hline 
B0 & A0: Faster RCNN $\rightarrow$ YOLOv4 & 81.7 & 56.6 & 44.5 & 60.9 \\ 
B1 & A0: Faster RCNN $\rightarrow$ YOLOX  & 83.5 & 57.2 & 44.8 & 61.8 \\
\hline
C0 & B1: w/ background change             & 83.7 & 57.5 & 44.9 & 62.0 \\
C1 & C0: w/ color augmentation            & 83.8 & 57.4 & 45.2 & 62.1 \\
\hline 
D0 & C1: ResNet-34 $\rightarrow$ ResNeSt-50d   & 84.7 & 60.0 & 47.3 & 64.0 \\
D1 & C1: ResNet-34 $\rightarrow$ ConvNeXt-base & 86.3 & 62.7 & 49.4 & 66.1 \\
D2 & D1: w/ amodal mask                        & 86.6 & 65.7 & 51.2 & 67.8 \\ 
D3 & D2: w/ class-aware head                   & 87.5 & 66.5 & 51.8 & 68.6 \\  
D4 & D2: One model per object                  & \textbf{88.7} & \textbf{70.1} & \textbf{54.9} & \textbf{71.3}  \\ 
\end{tabular}
\end{table}

\begin{table}
\caption{
\label{tab:lmo_abla_refine}
{\bf Ablation on LM-O for refinement module.}
We report the Average Recall (\%) of the BOP metric.
}
\centering
\begin{threeparttable}[c]
\tablestyle{4pt}{1.2}
\begin{tabular}{c| cccc | c c c | c}
Row & Coor. & Mask & F. W. & Sym. &
$\text{MSPD}$ &
$\text{MSSD}$ &
$\text{VSD}$ &
$\text{AR}_\text{{\tiny BOP}}$  \\
\shline
Init. & - & - & - & - &  88.7 & 70.1 & 54.9 & 71.3 \\
\hline
A & \xmark & \xmark & \xmark & \xmark &  87.2 & 82.2 & 62.9 & 77.5 \\
B & \cmark & \xmark & \xmark & \xmark &  77.7 & 73.6 & 57.9 & 69.7 \\
C & \cmark & \cmark & \xmark & \xmark &  80.5 &  76.4 & 61.5 & 72.8 \\
D & \cmark & \cmark & \cmark & \xmark &  89.5 & 84.9 & 65.4 & 79.9 \\
E & \cmark & \xmark & \cmark & \cmark & 88.6 & 83.7 & 65.0 & 79.1 \\
\hline
F & \cmark & \cmark & \cmark & \cmark &  \textbf{90.0} & \textbf{85.2} & \textbf{66.4} & \textbf{80.5}\\ 
\end{tabular}
\begin{tablenotes}
\item[] The row \textbf{Init.} is the initial pose from GDRN.
\item[] \textbf{Coor.} denotes whether the coordinate feature is used in the refinement module. \textbf{Mask} denotes whether the coordinate map is masked before extracting the coordinate feature as in Eq.~\ref{eq:coor_feat}. \textbf{F.W.} denotes whether the coordinate feature is weighted as in Eq.~\ref{eq:weight}. \textbf{Sym.} denotes whether the initial pose of the symmetric object is selected as in Eq.~\ref{eq:sym_pose}.
\end{tablenotes}
\end{threeparttable}
\end{table}
\begin{table}
\caption{
\label{tab:lmo_sota_refine}
\Rtwo{{\bf Comparison with other refinement methods on LM-O.}
We report the Average Recall (\%) of the BOP metric.}
}
\centering
\begin{threeparttable}[c]
\tablestyle{5.5pt}{1.2}
\begin{tabular}{c| c| c c c | c}
\rowcolor{RtwoColorBG} Method & Modality & 
$\text{MSPD}$ &
$\text{MSSD}$ &
$\text{VSD}$ &
$\text{AR}_\text{{\tiny BOP}}$  \\
\shline
GDRN (Init.) & RGB &  88.7 & 70.1 & 54.9 & 71.3 \\
\hline
CosyPose~\cite{labbe2020CosyPose}  & RGB & 86.8 & 66.7 & 52.2 & 68.5 \\
ICP~\cite{ICP}  & D  & 76.1 & 70.9 & 53.0 & 66.7 \\
FoundationPose~\cite{foundationposewen2024} & RGB-D &  86.0 &  82.0 & 63.7 & 77.2 \\
Ours & RGB-D & \textbf{90.0} & \textbf{85.2} & \textbf{66.4} & \textbf{80.5} \\
\end{tabular}
\end{threeparttable}
\end{table}

\begin{table*}[t]
    \centering
    \caption{
        {\bf Comparison with State of the Arts on the seven BOP core datasets.}
        We report the Average Recall (\%) of the BOP metric.
        The results for other methods are obtained from \href{https://bop.felk.cvut.cz/leaderboards/}{https://bop.felk.cvut.cz/leaderboards/}.
        For each column, we denote the best score in \textbf{bold} and the second best score in \textit{italics}.
        GDRNPP (BOP22) is the BOP Challenge 2022 version of GDRNPP, which utilizes \cite{lipson2022coupled} for depth refinement.
        \Rtwo{Compared to GDRNPP (BOP23), \ie~GPose2023 in the leaderboard, which utilizes YOLOv8~\cite{yolov8} as its detector, GDRNPP (YOLOX) employs YOLOX~\cite{ge2021yolox} for detection.}~
        \Rone{S.M. denotes if the method trains a single model for all objects on each dataset.}
    }
    \begin{threeparttable}[c]
    \tablestyle{6pt}{1.2}
    \small
    \begin{tabular}{l |cc >{\columncolor{RoneColorBG}}c|ccccccc |c | >{\columncolor{RtwoColorBG}}c}
        Method & Modality & Real & S.M.
        & LM-O & T-LESS & TUD-L & IC-BIN & ITODD & HB & YCB-V & Avg & time(s)
        \\ 
        \shline 
        \textbf{GDRNPP (Ours)} & RGB & \xmark & \xmark & 71.3 & \textbf{79.6} & \textbf{75.2} & \textbf{62.3} & \textbf{44.8} & \textit{86.9} & \textbf{71.3} & \textbf{70.2} & 0.28 \\
        PFA \cite{hu2022perspective} & RGB & \xmark & \cmark & \textbf{74.5} & 71.9 & \textit{73.2} & \textit{60.0} & 35.3 & 84.1 & 64.8 & 66.3 & 3.50 \\
        ZebraPose \cite{su2022zebrapose} & RGB & \xmark & \xmark & \textit{72.1} & 72.3 & 71.7 & 54.5 & \textit{41.0} & \textbf{88.2} & \textit{69.1} & \textit{67.0} & - \\
        SurfEmb \cite{haugaard2022surfemb} & RGB & \xmark & \cmark & 65.6 & \textit{74.1} & 71.5 & 58.5 & 38.7 & 79.3 & 65.3 & 64.7 & 8.89 \\
        EPOS \cite{hodan2020epos} & RGB & \xmark & \cmark & 54.7 & 46.7 & 55.8 & 36.3 & 18.6 & 58.0 & 49.9 & 45.7 & 	1.87 \\
        CDPNv2 \cite{li2019cdpn} & RGB & \xmark & \xmark & 62.4 & 40.7 & 58.8 & 47.3 & 10.2 & 72.2 & 39.0 & 47.2 & 0.98 \\
        DPODv2 \cite{shugurov2021dpodv2} & RGB & \xmark & \xmark & 58.4 & 63.6 & - & - & - & 72.5 & - & - & - \\ 
        CosyPose \cite{labbe2020CosyPose} & RGB & \xmark & \xmark & 63.3 & 64.0 & 68.5 & 58.3 & 21.6 & 65.6 & 57.4 & 57.0 & 0.48 \\ 
        \hline
        \textbf{GDRNPP (Ours)} & RGB & \cmark & \xmark & 71.3 & \textit{78.6} & 83.1 & \textbf{62.3} & \textbf{44.8} & \textit{86.9} & \textit{82.5} & \textbf{72.8} & 0.23 \\
        GDRNPP (S.M.) & RGB & \cmark & \cmark & 68.6 & 77.6 & 82.7 & \textit{61.7} & 26.0 & 80.9 & 76.8 & 67.8 & 0.23  \\
        PFA \cite{hu2022perspective} & RGB & \cmark & \cmark & \textbf{74.5} & 77.8 & \textit{83.9} & 60.0 & 35.3 & 84.1 & 80.6 & 70.9 & 3.02 \\
        ZebraPose \cite{su2022zebrapose} & RGB & \cmark & \xmark & \textit{72.1} & \textbf{80.6} & \textbf{85.0} & 54.5 & \textit{41.0} & \textbf{88.2} & \textbf{83.0} & \textit{72.0} & 0.25 \\
        SurfEmb \cite{haugaard2022surfemb} & RGB & \cmark &  \cmark  & 65.6 & 77.0 & 80.5 & 58.5 & 38.7 & 79.3 & 71.8 & 67.3 & 	8.89 \\
        CRT-6D \cite{castro2023crt} & RGB & \cmark &  \cmark  & 66.0 & 64.4 & 78.9 & 53.7 & 20.8 & 60.3 & 75.2  & 59.9 & 0.06 \\
        Pix2Pose \cite{park2019pix2pose} & RGB & \cmark &  \xmark  & 36.3 & 34.4 & 42.0 & 22.6 & 13.4 & 44.6 & 45.7 & 34.2 & 1.22 \\
        CDPNv2 \cite{li2019cdpn} & RGB & \cmark & \xmark & 62.4 & 47.8 & 77.2 & 47.3 & 10.2 & 72.2 & 53.2 & 52.9 & 	0.94 \\
        CosyPose \cite{labbe2020CosyPose} & RGB & \cmark & \cmark & 63.3 & 72.8 & 82.3 & 58.3 & 21.6 & 65.6 & 82.1 & 63.7 & 0.45 \\ 
        \hline
        \rowcolor{RtwoColorBG} \textbf{GDRNPP (BOP23)} & RGB-D & \xmark & \xmark & 79.4 & \textbf{89.0} & \textit{93.1} & \textbf{73.7} & \textbf{ 70.4} & \textbf{95.0} & 90.1 & \textbf{84.4} & 2.69 \\
        GDRNPP (YOLOX) & RGB-D & \xmark & \xmark & \textbf{80.5} & \textit{88.4} & 92.7 & \textit{73.4} & \textit{68.7} & \textit{94.4} &  \textbf{91.0}  & \textit{84.2} & 4.58 \\
        GDRNPP (BOP22) & RGB-D & \xmark & \xmark & 77.5 & 85.2 & 92.9 & 72.2 & 67.9 & 92.6 & \textit{90.6}  & {82.7} & 6.26 \\
        PFA \cite{hu2022perspective} & RGB-D & \xmark & \cmark &  \textit{79.7} & 80.2 & 89.3 & 67.6 & 46.9 & 86.9 & 82.6 & 76.2 & 2.63 \\
        SurfEmb \cite{haugaard2022surfemb} & RGB-D & \xmark & \cmark & 75.8 & 82.8 & 85.4 & 65.6 & 49.8 & 86.7 & 80.6 & 75.2 & 9.05 \\
        RCVPose3D \cite{wu2022keypoint} & RGB-D & \xmark & \cmark & 72.9 & 70.8 & \textbf{96.6} & 73.3 & 53.6 & 86.3 & 84.3 & 76.8 & 1.34  \\
        Drost \cite{drost2010model} & RGB-D & * & - & 51.5 & 50.0 & 85.1 & 36.8 & 57.0 & 67.1 & 37.5 & 55.0 & 87.57 \\ 
        Vidal Sensors \cite{vidal2018method} & D & * & - & 58.2 & 53.8 & 87.6 & 39.3 & 43.5 & 70.6 & 45.0 & 56.9 & 3.22 \\ 
        \hline
        \rowcolor{RtwoColorBG} \textbf{GDRNPP (BOP23)} & RGB-D & \cmark & \xmark & {79.4} & \textbf{91.4} & 96.4 & \textbf{73.7} & \textbf{70.4 }& \textbf{95.0} & \textit{92.8} & \textbf{85.6} & 2.67 \\
        GDRNPP (YOLOX) & RGB-D & \cmark & \xmark & \textbf{80.5} & \textit{89.5} & \textit{96.6} & \textit{73.4} & \textit{68.7} & \textit{94.4} & \textbf{92.9}  & \textit{85.1} & 4.58 \\
        GDRNPP (BOP22) & RGB-D & \cmark & \xmark & 77.5 & {87.4} & \textit{96.6} & {72.2} & {67.9} & {92.6} & {92.1}  & {83.7} & 6.26 \\
        PFA \cite{hu2022perspective} & RGB-D & \cmark & \cmark & \textit{79.7} & 85.0 & 96.0 & 67.6 & 46.9 & 86.9 & 88.8 & 78.7 & 2.32 \\
        ZebraPose \cite{su2022zebrapose} & RGB-D & \cmark & \xmark & 75.2 & 72.7 & 94.8 & 65.2 & 52.7 & 88.3 & 86.6 & 76.5 & 0.50 \\
        SurfEmb \cite{haugaard2022surfemb} & RGB-D & \cmark & \cmark & 75.8 & 83.3 & 93.3 & 65.6 & 49.8 & 86.7 & 82.4 & 76.7 & 9.05 \\
        CIR \cite{lipson2022coupled} & RGB-D & \cmark & \cmark & 73.4 & 77.6 & \textbf{96.8} & 67.6 & 38.1 & 75.7 & 89.3 & 74.1 & - \\
        CosyPose \cite{labbe2020CosyPose} & RGB-D & \cmark & \cmark & 71.4 & 70.1 & 93.9 & 64.7 & 31.3 & 71.2 & 86.1 & 69.8 & 13.74 \\ 
        Koenig-Hybrid \cite{konig2020hybrid} & RGB-D & \cmark & \cmark & 63.1 & 65.5 & 92.0 & 43.0 & 48.3 & 65.1 & 70.1 & 63.9 & 0.63 \\
        Pix2Pose \cite{park2019pix2pose} & RGB-D & \cmark & \xmark  & 58.8 & 51.2 & 82.0 & 39.0 & 35.1 & 69.5 & 78.0 & 59.1 &  4.84 \\
    \end{tabular}
    \medskip
    \begin{tablenotes} 
    \item[] ``Real'' means whether the method uses real-world data for training on T-LESS, TUD-L and YCB-V datasets.
    \item[] ``-'' denotes the results are unavailable and ``*'' denotes the method does not use the provided images for training.
    \end{tablenotes}
\end{threeparttable}
\label{tab:bop_sota}
\end{table*}

\subsection{Ablation Study on LM-O}

The BOP Challenge~\cite{hodan2018bop,hodan_bop20} has recently become the de-facto benchmark in object pose estimation. 
Therefore, to enhance our baseline method (TABLE~\ref{tab:lm_ablate} B0) for the BOP setup, we make several improvements and present the ablative results on the LM-O dataset in TABLE~\ref{tab:lmo_abla} and TABLE~\ref{tab:lmo_abla_refine}.

\noindent \textbf{Effectiveness of Detection.}
Due to the decoupling of the detector and pose estimator in our method, we can leverage the state-of-the-art detectors without re-training the network.
As a result, we evaluate GDRN with more recently developed detectors such as YOLOv4 \cite{bochkovskiy2020yolov4} and YOLOX \cite{ge2021yolox}. 
The results presented in TABLE~\ref{tab:lmo_abla} (B0, B1) demonstrate that the pose accuracy can be further enhanced by utilizing these more powerful detectors.

\noindent \textbf{Effectiveness of Image Augmentation.}
Considering that only synthetic data are available during training on the LM-O dataset, image augmentation plays a vital role in enhancing the generalization capability of object pose estimation methods, as demonstrated in \cite{labbe2020CosyPose,sundermeyer2018implicit}. 
During the training process, for each image, we randomly change the background to an image selected from the VOC dataset~\cite{Everingham10} with a probability of 0.5 (TABLE \ref{tab:lmo_abla} C0).
Additionally, color augmentation techniques, including dropout, Gaussian blur, Gaussian noise, and sharpness enhancement, are applied to augment 80\,\% of the images in the training phrase following~\cite{labbe2020CosyPose,sundermeyer2018implicit} (TABLE \ref{tab:lmo_abla} C1). 

\noindent \textbf{Ablation on Network Architecture.}
With the rapid growth of data amount (15,375 on LM \vs 349,693 on LM-O), GDRN needs a more powerful backbone with more parameters to increase the model's capacity.
TABLE~\ref{tab:lmo_abla} (D0, D1) shows that ResNeSt~\cite{zhang2022resnest} and ConvNeXt~\cite{liu2022convnet} outperform the basic ResNet~\cite{he2016deep} by a large margin.
Moreover, TABLE~\ref{tab:lmo_abla} (D2 \vs D1) reveals that predicting the amodal mask can effectively assist the network in dealing with occlusions, as mentioned in Section~\ref{sec:GDRN}.

We experiment with two class-ware settings and present the results in TABLE~\ref{tab:lmo_abla} (D3, D4).
Specifically, we first attempt to modify the output of the geometric head in a class-aware manner, where different object classes are assigned to individual output channels. 
This strategy allows the network to capture object-specific information, resulting in a noticeable performance improvement (68.6\,\% \vs 67.8\,\%).
Additionally, we conduct experiments by training a separate model for each object, which surpasses all previous results, achieving a remarkable performance of 71.3\,\% \wrt $\text{AR}_\text{{\tiny BOP}}$ metric leveraging pure RGB data. 

\noindent \textbf{Ablation on Refinement Module.}
The ablation on the refinement module is listed in TABLE \ref{tab:lmo_abla_refine}.
Without the coordinate map as input, the baseline (TABLE \ref{tab:lmo_abla_refine} A) improves the average recall from 71.3\,\% to 77.5\,\%.
As shown in TABLE \ref{tab:lmo_abla_refine} (B, C), %
by solely integrating the coordinate feature,
the performance drops significantly due to the erroneous coordinate map prediction.
However, by adding feature weighting, the average recall reaches 79.9\,\%, which is 2.1\,\% higher than the baseline (TABLE \ref{tab:lmo_abla_refine} D \vs A).
It reveals that feature weighting is essential in improving the robustness against error in the input coordinate and preventing performance degradation.
By comparing TABLE \ref{tab:lmo_abla_refine} (D, F), it can be seen that selecting a proper initial pose of the symmetric objects as in Eq.~\ref{eq:sym_pose} brings 0.6\,\% performance gain.
TABLE~\ref{tab:lmo_abla} (E \vs F) proves that masking the input coordinate map as in Eq.~\ref{eq:coor_feat} is also important since it filters out the outliers dynamically.

\begin{figure}[t]
    \begin{center}
    \includegraphics[width=0.96\linewidth]{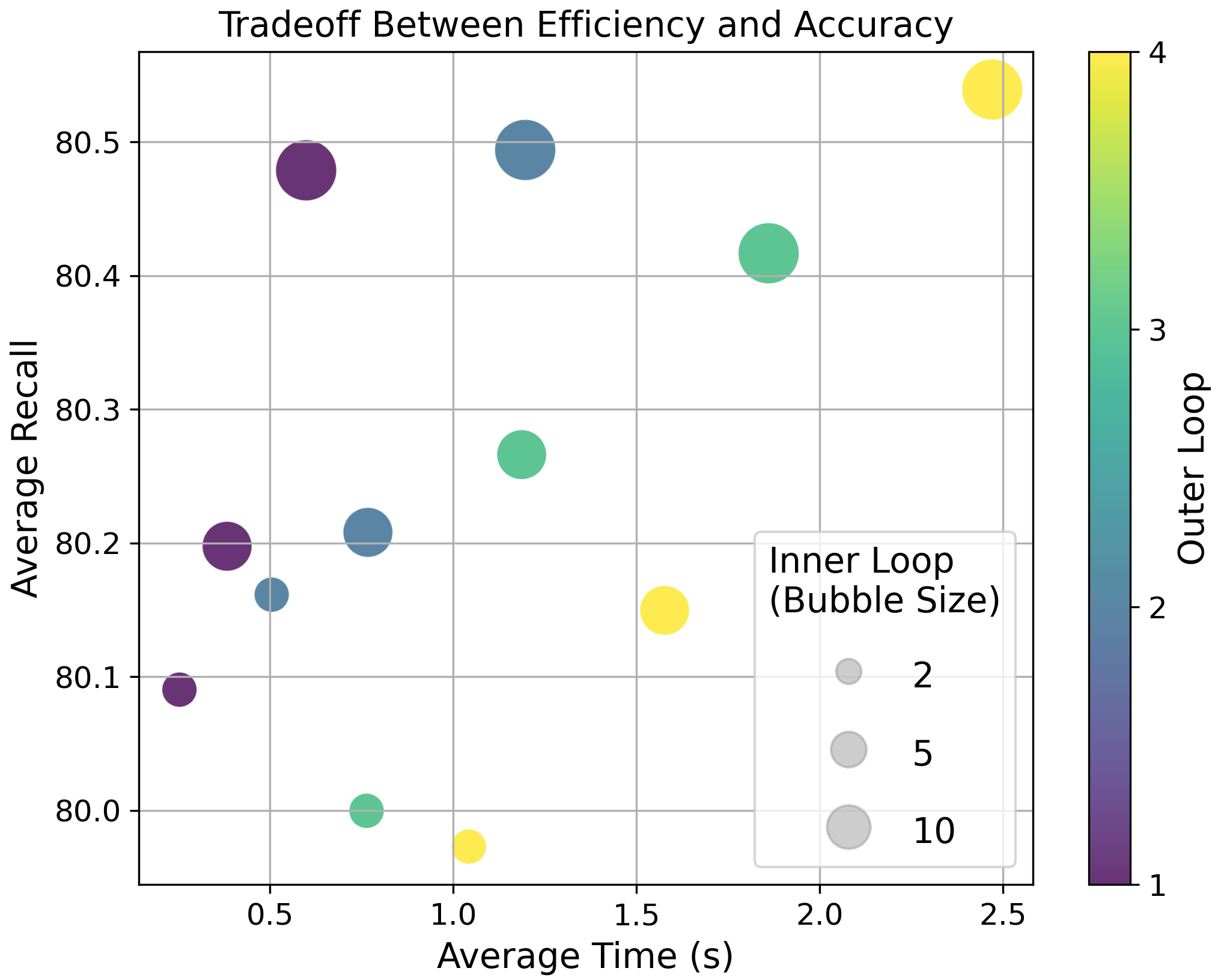}
    \end{center}
    \caption{
    \label{fig:efficiency_ablation}
    \Rtwo{
    {\bf Efficiency \vs accuracy with varying inner and outer loop iterations for the refinement module on LM-O.}
    The bubble size represents the inner loop number, while the color indicates the outer loop number.
}
}
\end{figure}

\Rtwo{
Fig. \ref{fig:efficiency_ablation} illustrates the trade-off between efficiency and accuracy \wrt the refinement module.
When the inner loop number ($T$ defined in Sec. \ref{sec:refine}) is set to 2 and the outer loop number ($N_\text{out}$ in Sec. \ref{sec:refine}) to 1, the average recall decreases by 0.5\%, while the inference time drops significantly from 2.48 s to 0.25 s.
The optimal values for the inner and outer loop numbers can be selected based on the specific requirements of the real-world application.
}

\begin{figure*}[htbp]
    \begin{center}
    \includegraphics[width=0.95\linewidth]{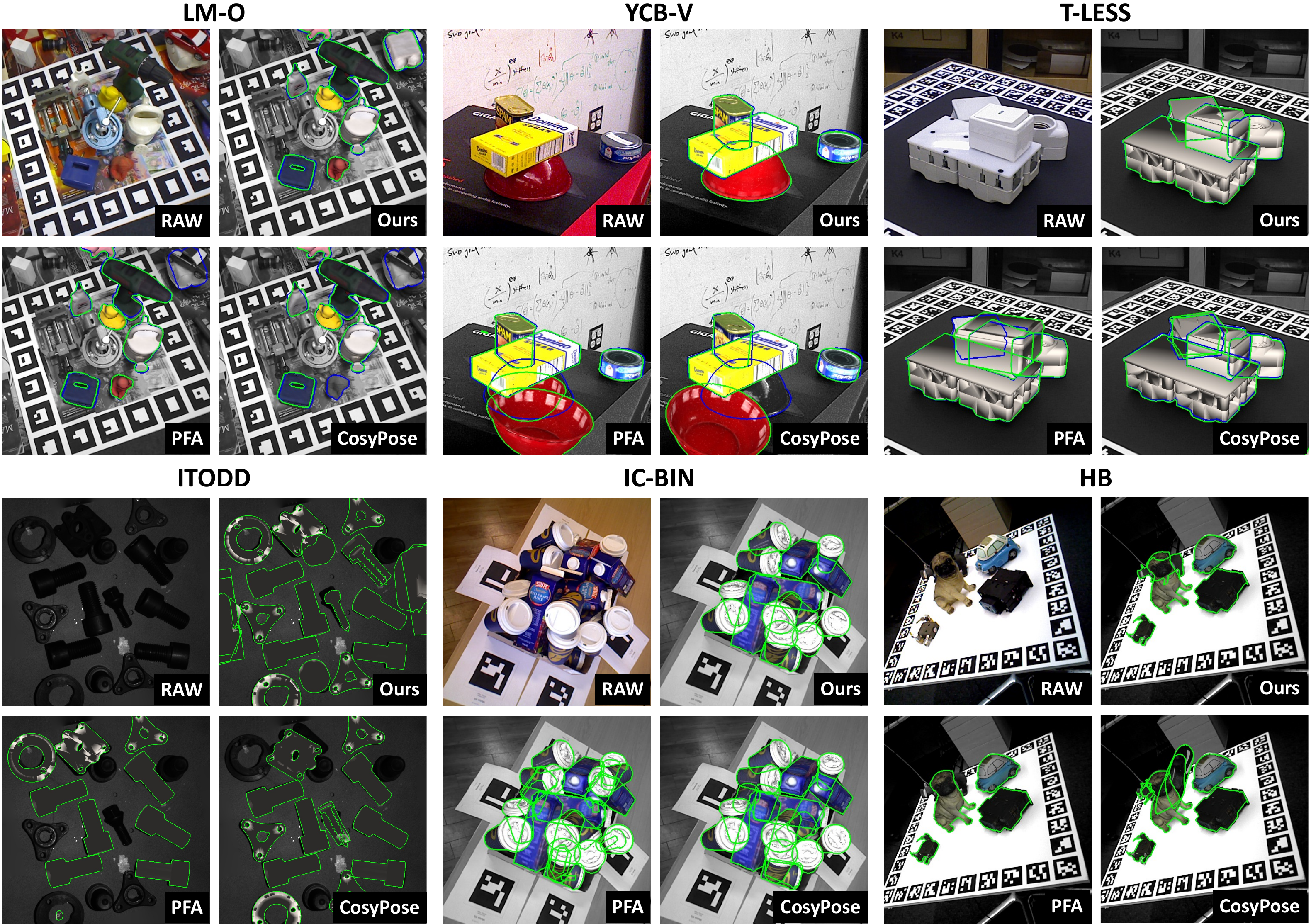}
    \end{center}
    \caption{
        \label{fig:exp_vis}
        {\bf Qualitative results on six datasets.}
        We compare our method with PFA~\cite{hu2022perspective} and CosyPose~\cite{labbe2020CosyPose}, maintaining a consistent experimental setup using depth and real images.
        For each image, we visualize the predicted 6D poses by rendering the 3D models and overlaying them onto the grayscale image.
        Predicted poses are demonstrated in \emph{Green} contours and ground-truth poses are demonstrated in \emph{Blue} contours (if have).
    }
\end{figure*}

\subsection{Comparison with State of the Arts}
\Rtwo{We compare our depth refinement module with several state-of-the-art refinement methods~\cite{ICP,labbe2020CosyPose,foundationposewen2024}, and present the results in TABLE~\ref{tab:lmo_sota_refine}. 
As shown in the table,
our proposed geometry-guided depth refinement method outperforms all other methods, achieving the highest accuracy. 
The results also indicate that the performance of ICP~\cite{ICP} and CosyPose~\cite{labbe2020CosyPose} shows a slight decline compared to the initial predictions.
The reliance on a single modality for refinement, \ie~CosyPose using RGB and ICP using only depth, constrains their performance.
Notably, the novel object pose refinement method FoundationPose~\cite{foundationposewen2024} achieves a performance closest to ours.
}

TABLE~\ref{tab:bop_sota} compares our enhanced approach (GDRNPP) with state-of-the-art methods on the seven core datasets included in the BOP benchmark. 
Remarkably, GDRNPP significantly outperforms all other state-of-the-art methods like PFA~\cite{hu2022perspective} ZebraPose~\cite{su2022zebrapose}, SurfEmb~\cite{haugaard2022surfemb}, CPDNv2~\cite{li2019cdpn}, CosyPose~\cite{labbe2020CosyPose}, CIR~\cite{lipson2022coupled}, and RCVPose3D~\cite{wu2022keypoint} across various data modalities (RGB and RGB-D) and domains (synthetic and real).
Specifically, utilizing only synthetic RGB data for training, our method achieves an average recall of 70.2\,\% \wrt the $\text{AR}_\text{{\tiny BOP}}$ metric, exceeding the second top-performing method ZebraPose~\cite{su2022zebrapose} by 3.2\,\%. 
Furthermore, when real data is available on T-LESS, TUD-L, and YCB-V datasets, the performance increases to 72.8\,\% without any refinement.
Our single model for each dataset (67.8\,\%) is also comparable with other methods.
Noteworthy, our pure RGB-based method even surpasses the RGB-D based method CosyPose relying on ICP for refinement (72.8\,\% \vs 69.8\,\%), which is the previously top-performing method in the BOP 2020 Challenge~\cite{hodan_bop20}.

Utilizing RGB-D images, our method achieves an average recall of 85.6\,\% with real data and 84.4\,\%  with only synthetic data.
The BOP22 version of GDRNPP, incorporating \cite{lipson2022coupled} for pose refinement, significantly outperforms other competitors and wins ``The Overall Best Method'' of the BOP 2022 Challenge~\cite{sundermeyer2023bop}.
\Rtwo{By adopting the geometry-guided pose refinement module and a more powerful detector~\cite{yolov8}, the average recall further improves upon~\cite{lipson2022coupled} by 1.9\,\% with real data and 1.7\,\% without real data, winning us ``The Overall Best Method'' of the BOP 2023 Challenge~\cite{hodan2024bop}.
Remarkably, the current version of GDRNPP achieves state-of-the-art performance on five out of the seven BOP core datasets.}

\Rtwo{We highlight the effect of the detector by comparing YOLOX~\cite{ge2021yolox} and YOLOv8~\cite{yolov8} and present the results in TABLE~\ref{tab:bop_sota}.
Even with YOLOX as the detector, GDRNPP still exhibits competitive results on the BOP benchmark, which shows the robustness of the pose estimator.
Generally, a more accurate detector would lead to more precise pose estimation (YOLOv8 85.6~\% \vs YOLOX 85.1~\% with real data).
Nevertheless, the prominent improvements of GDRNPP are in the enhancements to the pose estimator and refiner parts rather than the stronger detector.}

Fig.~\ref{fig:exp_vis} illustrates some additional qualitative results for LM-O, YCB-V, T-LESS, ITODD, IC-BIN, and HB.
Compared to PFA~\cite{hu2022perspective} and CosyPose~\cite{labbe2020CosyPose}, GDRNPP shows superior performance with fewer missing and falsely detected objects, while also producing more precise pose estimations.
Notably, GDRNPP also demonstrates its versatility in intricate scenarios exhibiting clutter, occlusion, and varying lighting conditions.

\subsection{Runtime Analysis}
\begin{figure}
\centering
 \begin{tikzpicture}
    \hspace{-4pt}
    \begin{axis}[xlabel=Runtime(s),ylabel=$\text{AR}_\text{{\tiny BOP}}$,
    x label style={at={(axis description cs:0.98,0.00)},anchor=north},
    y label style={at={(axis description cs:-0.1,0.90)},anchor=west},
    ymin=20, ymax=81, legend pos=south east,
    ymajorgrids=true,
    axis x line*=bottom,
    axis y line*=left,
    width=0.9\columnwidth,
    height=0.8\columnwidth,
    every mark/.append style={transform shape}]
    \addplot [mark=text, text mark=\huge $\filledstar$, mark size=5, purple]
coordinates {(0.23, 72.8)};
    \addplot [mark=*, color=teal, mark size=2,mark options={fill=teal}]
    coordinates {
    (0.25,72.0)
    };
    \addplot [mark=*, color=teal, mark size=2,mark options={fill=teal}]
    coordinates {
    (3.02, 70.9)
    };
    \addplot [mark=*, color=gray, mark size=2,mark options={fill=gray}]
    coordinates {
    (8.89, 67.3)
    };
    \addplot [mark=*, color=brown, mark size=2,mark options={fill=brown}]
    coordinates {
    (0.06, 59.9)
    };
    \addplot [mark=*, color=green, mark size=2,mark options={fill=green}]
    coordinates {
    (1.22, 34.2)
    };
    \addplot [mark=*, color=yellow, mark size=2,mark options={fill=yellow}]
    coordinates {
    (0.94, 52.9)
    };
    \addplot [mark=*, color=orange, mark size=2,mark options={fill=orange}]
    coordinates {
    (0.45, 63.7)
    };

    \node at (axis cs:0.23,75.8) {\textbf{Ours}};
    \node at (axis cs:0.5,70.0) {ZebraPose};
    \node at (axis cs:3.02,73.9){PFA} ;
    \node at (axis cs:8.69,70.3) {SurfEmb};
    \node at (axis cs:1.20,59.9) {CRT-6D};
    \node at (axis cs:1.22,37.2) {Pix2Pose};
    \node at (axis cs:0.94,49.9) {CDPNv2};
    \node at (axis cs:0.45,66.7) {CosyPose};
    
    \end{axis}
\end{tikzpicture}
\begin{tikzpicture}
    \hspace{-4pt}
    \begin{axis}[xlabel=Runtime(s),ylabel=$\text{AR}_\text{{\tiny BOP}}$,
    x label style={at={(axis description cs:0.98,-0.05)},anchor=north},
    y label style={at={(axis description cs:-0.1,0.90)},anchor=west},
    ymin=50, ymax=90, legend pos=south east,
    xmin=0, xmax=15, legend pos=south east,
    ymajorgrids=true,
    axis x line*=bottom,
    axis y line*=left,
    width=0.9\columnwidth,
    height=0.8\columnwidth,
    every mark/.append style={transform shape}]
    \addplot [mark=text, text mark=\huge $\filledstar$, mark size=5, purple]
coordinates {(2.67,85.6)};
    \addplot [mark=*, color=teal, mark size=2,mark options={fill=teal}]
    coordinates {
    (0.50,76.5)
    };
    \addplot [mark=*, color=teal, mark size=2,mark options={fill=teal}]
    coordinates {
    (2.32, 78.7)
    };
    \addplot [mark=*, color=gray, mark size=2,mark options={fill=gray}]
    coordinates {
    (9.05, 76.7)
    };
    \addplot [mark=*, color=brown, mark size=2,mark options={fill=brown}]
    coordinates {
    (13.74, 69.8)
    };
    \addplot [mark=*, color=green, mark size=2,mark options={fill=green}]
    coordinates {
    (0.63, 63.9)
    };
    \addplot [mark=*, color=yellow, mark size=2,mark options={fill=yellow}]
    coordinates {
    (4.84, 59.1)
    };

    \node at (axis cs:3.0,87.5) {\textbf{Ours}};
    \node at (axis cs:2.0,74.8) {ZebraPose};
    \node at (axis cs:2.82,80.5){PFA} ;
    \node at (axis cs:9.5,78.5) {SurfEmb};
    \node at (axis cs:13.4,71.2) {CosyPose};
    \node at (axis cs:2.8,65.3) {Koenig-Hybrid};
    \node at (axis cs:5.2,61.0) {Pix2Pose};
    \end{axis}
\end{tikzpicture}
\caption{\Rtwo{\textbf{Runtime analysis under RGB (upper) and RGB-D (lower) modality using real data for training.}}
We report the Average Recall (\%) of BOP metric \wrt the average runtime (second) obtained from \href{https://bop.felk.cvut.cz/leaderboards/} {https://bop.felk.cvut.cz/leaderboards/}.
Results show that our method gains the highest score while maintaining a fast inference speed.
}
\label{fig:runtime}
\end{figure}
Fig.~\ref{fig:runtime} depicts the average runtime of our algorithm, along with current state-of-the-art methods in the BOP Challenge leaderboard.
We plot $\text{AR}_\text{{\tiny BOP}}$ (\%) versus inference time (second) to intuitively show the performance of each method trained with real-world data.

Compared with indirect methods which rely on 2D-3D or 3D-3D correspondence like ~\cite{li2019cdpn,park2019pix2pose},
our method offers a compelling combination of real-time performance and accurate pose estimation. 
This achievement is attributed to our fully learning-based strategy, eliminating the time-consuming and inaccurate P$n$P/RANSAC procedure.
Specifically, GDRN runs at the average speed of 0.23s per RGB image, gains 97\,\% and 92\,\% leap forward against SurfEmb \cite{haugaard2022surfemb} (8.89s) and PFA \cite{hu2022perspective} (3.02s) respectively, which are the two most competitive methods towards GDRN \wrt the BOP metric.
\Rtwo{When considering depth refinement, GDRNPP runs slightly slower at 2.67s, but achieves significantly higher accuracy at 85.6\%.
Compared to other methods with faster inference speeds like ZebraPose (0.5s), Koenig-Hybrid (0.63s), and PFA (2.32s), GDRNPP excels in terms of pose estimation accuracy.
}

\section{Conclusion}
In this work, we have proposed a geometry-guided and fully learning-based pose estimator to eliminate the drawbacks of indirect pipelines.
To directly regress 6D poses from monocular images, we exploit the intermediate geometric features regarding 2D-3D correspondences organized regularly as image-like 2D patches,
and utilize a learnable 2D convolutional Patch-P$n$P to replace the P$n$P/RANSAC stage. 
Furthermore, we harness depth to refine the pose by establishing 3D-3D dense correspondences between observed and rendered RGB-D images.
With geometric guidance, the network dynamically removes outliers, thereby enabling us to solve the pose in a differentiable fashion.
Our fully learning-based pipeline shows competitive performance in various challenging scenarios while maintaining a fast inference speed.
In the future, we want to extend our work to more challenging scenarios, such as the lack of annotated real data~\cite{wang2021occlusion} and
unseen object categories or instances~\cite{wang2019nocs,park2020latentfusion}.

\section*{Acknowledgements}
This work was supported in part by the National Natural Science Foundation of China under Grant No. 62406169, 
and in part by the China Postdoctoral Science Foundation under Grant No. 2024M761673.

{\small
\bibliographystyle{IEEEtran}
\bibliography{main}
}

\begin{IEEEbiography}[{\includegraphics[width=1in,height=1.25in,clip,keepaspectratio]{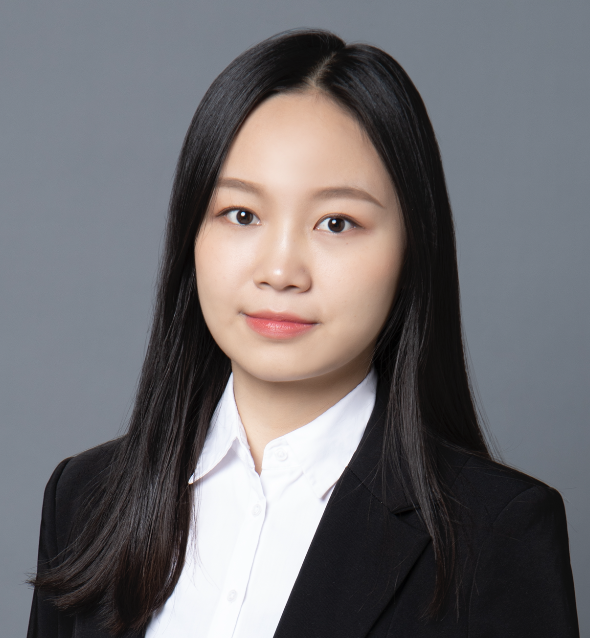}}]{Xingyu Liu}
is currently a Ph.D. student in the Department of Automation, at Tsinghua University supervised by Xiangyang Ji. She received her B.E. degree from the Department of Automation, Beihang University, Beijing, China, in 2021.
Her research interests lie in 3D computer vision and robotic vision.
\end{IEEEbiography}

\begin{IEEEbiography}[{\includegraphics[width=1in,height=1.25in,clip,keepaspectratio]{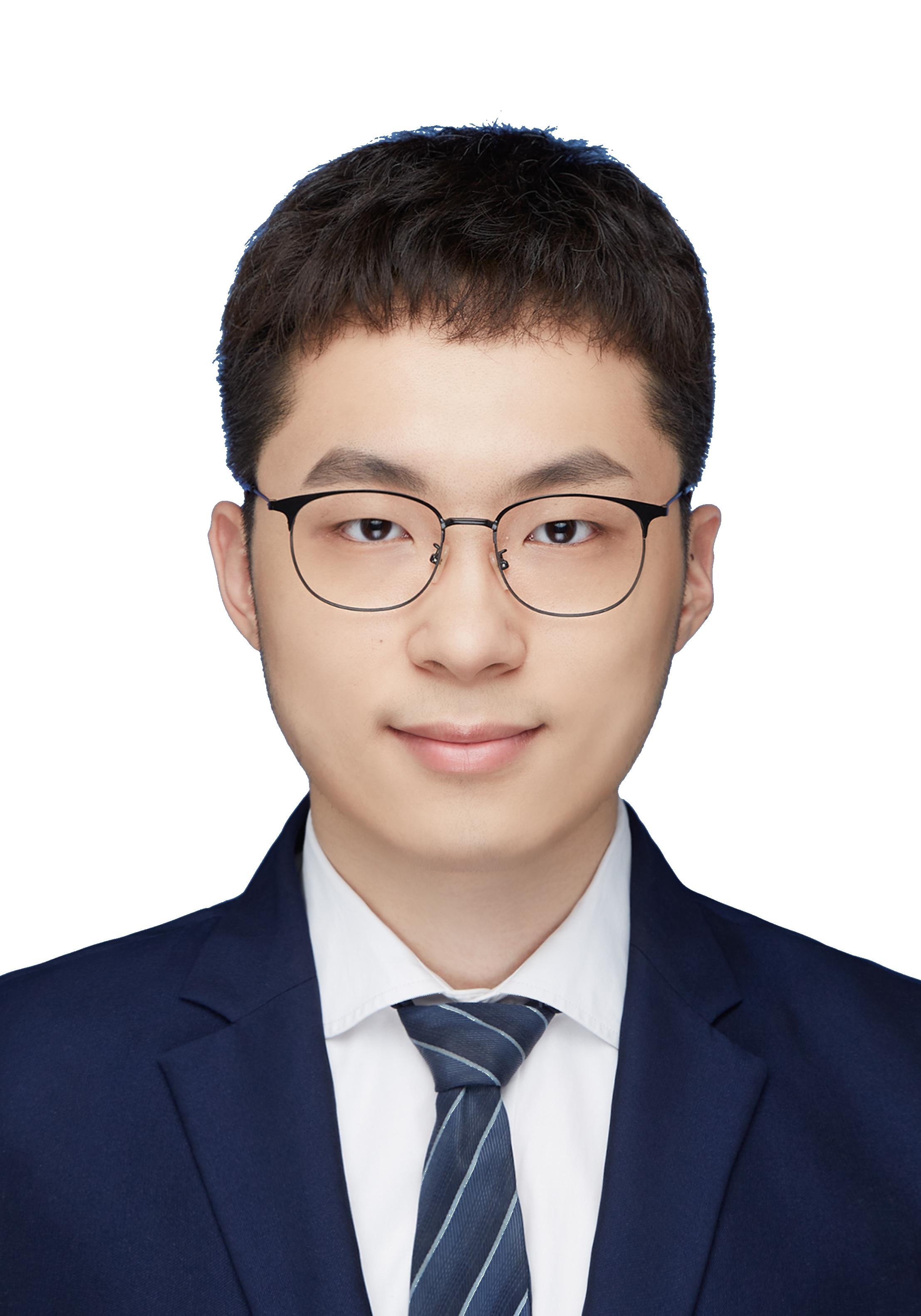}}]{Ruida Zhang}
received the B.E. degree from the Department of Automation, Tsinghua University, Beijing, China, in 2021. He is working toward a Ph.D. degree at Tsinghua University, under the supervision of Xiangyang Ji. His research interests include 3D computer vision and robotics.
\end{IEEEbiography}

\begin{IEEEbiography}[{\includegraphics[width=1in,height=1.25in,clip,keepaspectratio]{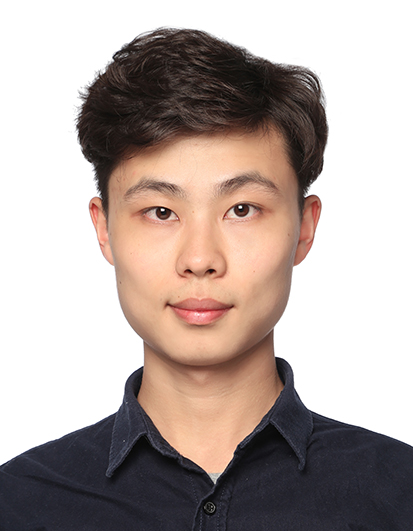}}]
{Chenyangguang Zhang} is currently a M.S. student in the Department of Automation, at Tsinghua University, supervised by Xiangyang Ji. He received a B.E. degree from the Department of Automation, Tsinghua University, Beijing, China, in 2022. His research interests lie in 3D computer vision and deep learning.
\end{IEEEbiography}

\begin{IEEEbiography}[{\includegraphics[width=1in,height=1.25in,clip,keepaspectratio]{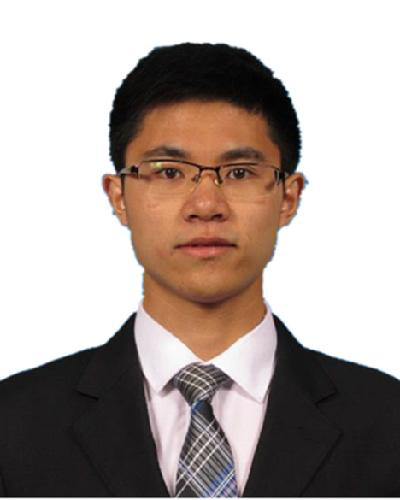}}]{Gu Wang} received B.E. and Ph.D. degrees from Department of Automation, Tsinghua University, Beijing, China, in 2016 and 2022, respectively. He was a visiting scholar at Technical University of Munich from 2019 to 2020. He was a Doctoral Management Trainee at JD.com from 2022 to 2023, working on calibration, localization and mapping in autonomous driving. 
He is currently a postdoctoral researcher at Tsinghua University, Beijing. His research interests include 3D computer vision, and vision in robotics.
\end{IEEEbiography}

\begin{IEEEbiography}[{\includegraphics[width=1in,height=1.25in,clip,keepaspectratio]{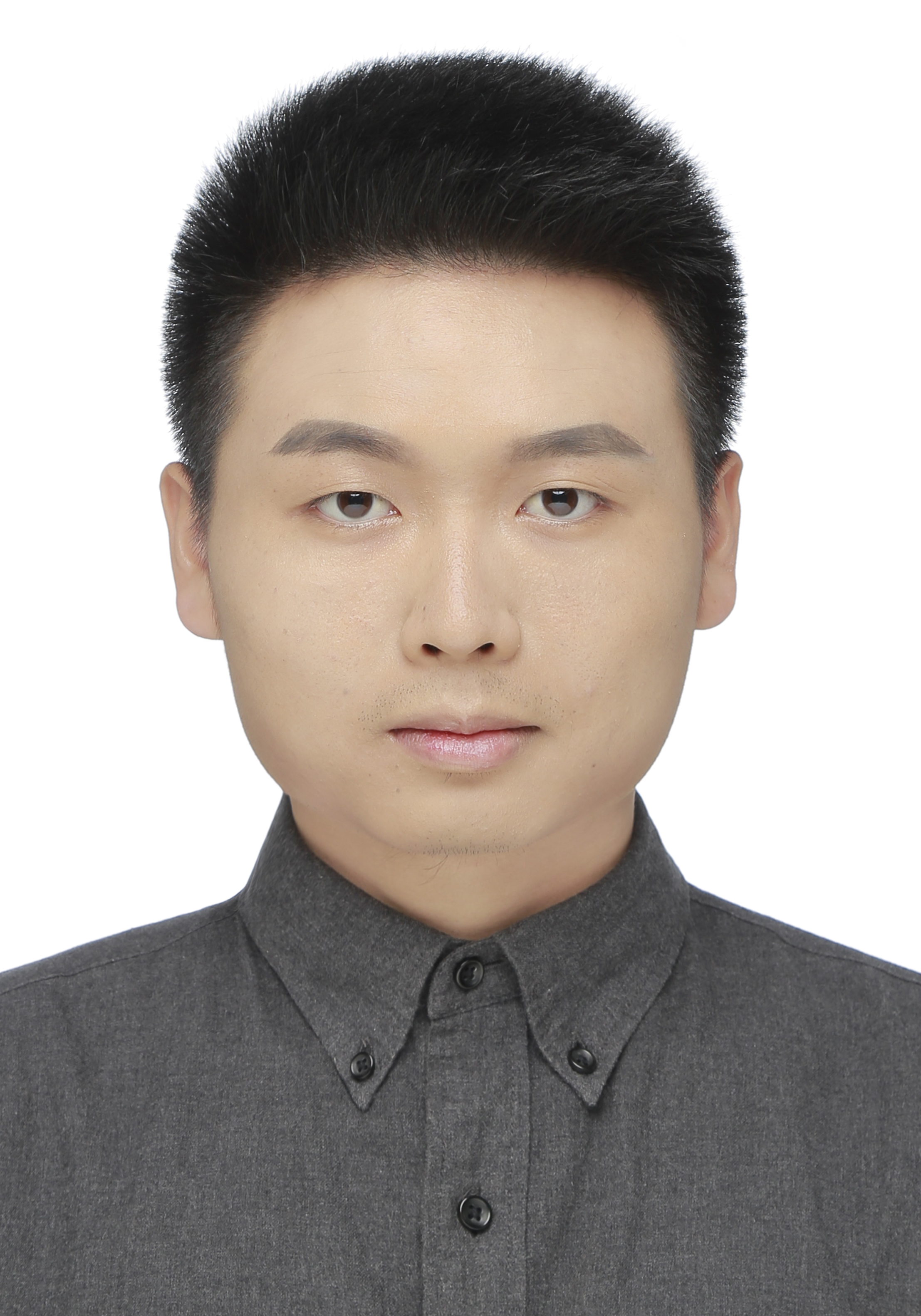}}]
{Jiwen Tang} received the B.E. degree from Wuhan University, Wuhan, China, in 2014 and the Ph.D. degree from Aerospace Information Research Institute, Chinese Academy of Sciences, Beijing, China, in 2021. From 2021 to 2024, he was a postdoctoral fellow with the Department of Automation, Tsinghua University, Beijing, China. In 2024, he joined China University of Geosciences Beijing, where he is currently a lecturer with the School of Information Engineering. His research interests lie in computer vision and deep learning.
\end{IEEEbiography}

\begin{IEEEbiography}[{\includegraphics[width=1in,height=1.25in,clip,keepaspectratio]{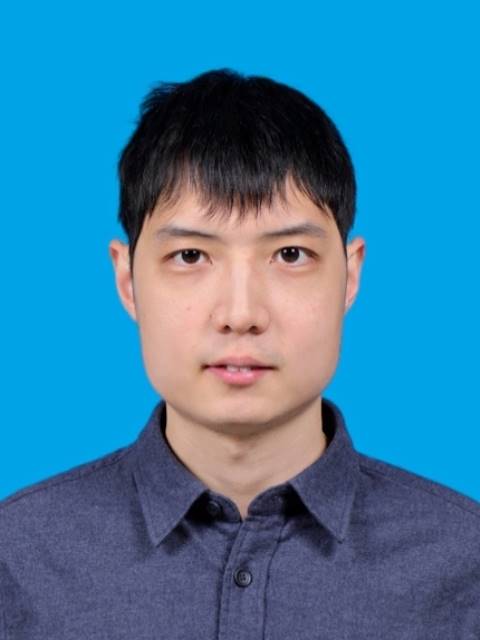}}]{Zhigang Li} received the B.E. degree in the School of Automation Science and Electrical Engineering from Beihang University, in 2015, and the PhD degree in the Department of Automation from Tsinghua University, in 2021. His research interests are computer vision, deep learning, and object pose estimation.
\end{IEEEbiography}

\begin{IEEEbiography}[{\includegraphics[width=1in,height=1.25in,clip,keepaspectratio]{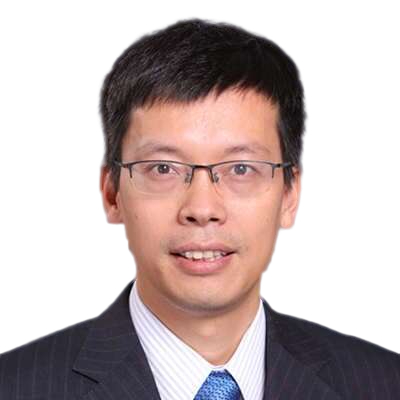}}]{Xiangyang Ji} received the B.E. degree in materials science and the M.S. degree in computer science from the Harbin Institute of Technology, Harbin, China, in 1999 and 2001, respectively, and the Ph.D. degree in computer science from the Institute of Computing Technology, Chinese Academy of Sciences, Beijing, China. He joined Tsinghua University, Beijing, in 2008, where he is currently a Professor at the Department of Automation, School of Information Science and Technology. 
He has authored more than 200 refereed conference and journal papers. His current research interests include signal processing, computer vision and computational photography.
\end{IEEEbiography}

\end{document}